\documentclass{article}

\usepackage[nonatbib,preprint]{neurips_2019}

\usepackage[utf8]{inputenc} % allow utf-8 input
\usepackage[T1]{fontenc}    % use 8-bit T1 fonts
\usepackage{hyperref}       % hyperlinks
\usepackage{url}            % simple URL typesetting
\usepackage{booktabs}       % professional-quality tables
\usepackage{amsfonts}       % blackboard math symbols
\usepackage{nicefrac}       % compact symbols for 1/2, etc.
\usepackage{microtype}      % microtypography

\usepackage[toc,page]{appendix}
\usepackage{todonotes}
\usepackage{microtype}

\usepackage{graphicx}
\usepackage{subfigure}
\usepackage{enumitem}
\usepackage{comment}
\usepackage{booktabs} % for professional tables
\graphicspath{{./images/}}
\DeclareGraphicsExtensions{.pdf,.jpeg,.jpg,.eps,.png}

\usepackage{amsmath, amsthm, amssymb, xspace}

\newcommand{\DNFP}{\emph{NFP}}
\newcommand{\FP}{\emph{NeuralFingerprinting}}
\newcommand{\FPdata}{\texttt{NFPdata}}
\newcommand{\fp}{\chi}
\newcommand{\fpx}{\Delta x}
\newcommand{\fpy}{\Delta y}
\newcommand{\fpxy}{\fp}
\newcommand{\fploss}{L_{\textrm{fp}}}
\newcommand{\pert}{\eta}

\newcommand{\INST}[1]{}
\newcommand{\KL}[1]{\textbf{KL}}

\usepackage{color}
\definecolor{blue}{rgb}{0,0,.7}
\definecolor{red}{rgb}{.7,0,0}
\definecolor{orange}{rgb}{1,.6,0}
\definecolor{purple}{rgb}{.4,0,.5}
\definecolor{brown}{rgb}{.4,.2,.1}
\definecolor{green}{rgb}{0,.5,0}

\newcommand{\refn}[1]{(\ref{#1})}

\newcommand{\brck}[1]{\left(#1\right)}

\newcommand{\brckcur}[1]{\left\{#1\right\}}

\newcommand{\brcka}[1]{\langle #1\rangle}
\newcommand{\norm}[1]{\|#1\|}

\newcommand{\fr}[2]{\frac{#1}{#2}}

\newcommand{\abs}[1]{\left|#1\right|}
\newcommand{\be}{\begin{equation}}
\newcommand{\ee}{\end{equation}}
\newcommand{\bali}{\begin{eqnarray*}}
	\newcommand{\eali}{\end{eqnarray*}}
\newcommand{\eq}[1]{\begin{align}#1\end{align}}

\newcommand{\calI}{\mathcal{I}}

\newcommand{\calO}{\mathcal{O}}

\newcommand{\sign}[1]{\textrm{sign }#1}

\usepackage{diagbox}
\usepackage{graphicx}
\usepackage[T1]{fontenc}    % use 8-bit T1 fonts
\usepackage[utf8]{inputenc} % allow utf-8 input
\usepackage[numbers]{natbib}
\usepackage{floatrow}
\usepackage{algorithm}
\usepackage{amsfonts}       % blackboard math symbols
\usepackage{amsmath}
\usepackage{amssymb}
\usepackage{amsthm}
\usepackage{bm}
\usepackage{booktabs}      % professional-quality tablesgu
\usepackage{enumitem}
\usepackage{color}
\usepackage{algorithmic}
\usepackage{comment}
\usepackage{multirow}
\usepackage{enumitem}
\usepackage{microtype}      % microtypography
\usepackage{nicefrac}       % compact symbols for 1/2, etc.
\usepackage{tabularx,ragged2e,booktabs,caption}
\usepackage{url}   
\usepackage{breakurl}
\usepackage{wrapfig}
\usepackage{sidecap}
\usepackage{hyperref}
\newtheorem{thm}{Theorem}

\DeclareMathOperator*{\argmax}{argmax}

\definecolor{mygreen}{rgb}{0,0.4,0}
\definecolor{myred}{rgb}{0.4,0,0}
\newcommand{\greencomment}[1]{\textit{\textcolor{mygreen}{\##1}}}
\newcommand{\redcomment}[1]{\textit{\textcolor{myred}{\##1}}}

\title{Detecting Adversarial Examples using Neural Fingerprinting}
\author{%
Sumanth Dathathri\thanks{Equal contribution.} \\
Caltech \\
sdathath@caltech.edu\\
\And 
Stephan Zheng$^{*}$\thanks{Work done while at Caltech.} \\
Salesforce Research \\
stephan.zheng@salesforce.com \\
\And
Tianwei Yin \\
University of Texas at Austin \\
yintianwei@utexas.edu \\
\And
Yisong Yue \\
Caltech \\
yyue@caltech.edu \\
\And 
Richard M. Murray \\
Caltech \\
murray@cds.caltech.edu 
}
\begin{document}

\maketitle

\begin{abstract}
	%EXTRASPACE
	Deep neural networks are vulnerable to adversarial examples: input data that has been manipulated to cause dramatic model output errors. 
	%EXTRASPACE
	%
	To address this challenge, we propose \FP{} (\DNFP{}): a simple, yet effective method to detect adversarial examples (in the grey-box and black-box threat models) for image data. \DNFP{} relies on verifying whether prediction behavior is consistent with a set of \textit{fingerprints}. These fingerprints are encoded into the prediction response around the data-distribution during training, and are inspired by biometric and cryptographic signatures, and we provide a theoretical characterization of \DNFP{} for linear networks.
	%EXTRASPACEs
	%
	%The benefits of our method are that 1) it is fast, 2) it is prohibitively expensive for an attacker to reverse-engineer which fingerprints were used, and 3) it does not assume knowledge of the adversary.
	%EXTRASPACE
	%In this work, we theoretically analyze \DNFP{} for linear networks, 
	We also show that \DNFP{} significantly improves on state-of-the-art adversarial detection mechanisms for deep neural networks, by detecting a full range of known adversarial attacks with \emph{98-100\%} AUC-ROC scores on the MNIST, CIFAR-10 and MiniImagenet (20 classes) datasets.
	%and 3) show that networks trained with \DNFP{} are able to significantly outperform state-of-the-art techniques on several outlier detection tasks.
	%EXTRASPACE
	We analyze \DNFP{} in detail under the grey/black-box threat models, when the adversary has limited or no-access to the exact fingerprints.
	%EXTRASPACE
	Further, in the white-box setting, we find that \DNFP{} is robust against a full-array of the strongest-known attacks that have full knowledge of the defense; we subsequently developed a new adaptive white-box attack to fool \DNFP{}.
\end{abstract}
%\vspace{-10pt}
\section{Introduction}\label{sec:intro}

Deep neural networks (DNNs) are highly effective pattern-recognition models for a wide range of tasks, e.g., computer vision \citep{objectrecog}, speech recognition \citep{speechrecog} and sequential decision-making \citep{PlayGo}.
%
%However, DNNs are typically vulnerable to making high confidence predictions on adversarial data not encountered during training \
However, DNNs are vulnerable to adversarial examples: an attacker can add (small) perturbations to input data that are imperceptible to humans, and can drastically
change the model’s output, introducing catastrophic errors \citep{YC15, szegedy2013intriguing, GoodfellowSS14}. 
When DNNs encounters adversarial data, one desirable response is to flag the data and abstain from making confident predictions, thereby enabling DNNs to function reliably in noisy  environments or mission-critical applications, e.g., in autonomous vehicles.

In this work,  we study the problem of detecting adversarial data  (after which the DNN abstains from making predictions). %, thereby allowing the network to abstain from making confident mistakes. 
We propose \FP{} (\DNFP): a fast and secure method to training DNNs to robustly detect \emph{grey}-box and \emph{black}-box adversarial attacks \citep{securitytesting}.
%
%In the adversarial case, we focus on detecting adversarial data in the \emph{grey}-box and \emph{black}-box threat models \citep{securitytesting}. 
%
The grey-box setting is motivated by the practical security setting where a private key is used to secure the system from adversaries \citep{security}.
For \DNFP{}, this specific key is the exact sequence of randomly
generated fingerprints.

The core idea of \DNFP{} is to encode fingerprint patterns into the response of a DNN around real data.
These patterns characterize the DNN's expected behavior around real data and can be used to detect adversarial examples.
This approach is attractive as encoding fingerprints is simple to implement during training, and evaluation is computationally inexpensive.
Furthermore, \DNFP{} is agnostic of the adversary's attack mechanism.  We provide a formal characterization of \DNFP{} for linear classification.

We conduct a thorough empirical evaluation of \DNFP{} across a range of adversarial attacks in computer vision.
We show that \DNFP{} achieves state-of-the-art near-perfect AUC-ROC scores on detecting and separating unseen test data and adversarial examples in the grey- and black-box threat models.
The performance of \DNFP{} is demonstrated to be robust to the choice of fingerprints and hyperparameters.
%\footnote{Code to reproduce experiments: {\tiny\url{https://www.dropbox.com/sh/jaf9rbfvr8lf9tt/AADFqfxEY3KSZyc0PNbBfATIa?dl=0}}.}
Finally, we find that \DNFP{} is robust to a full-range previously known white-box adversarial attacks; we subsequently developed a new adaptive white-box attack to break \DNFP{} (See Appendix \ref{sec:expadapt}).

%To summarize, our key contributions are:
%EXTRASPACE
%\begin{itemize}%[noitemsep]
%\item We present \DNFP{}: a simple and secure method to reliably detect adversarial data
%\item We formally characterize the effectiveness of \DNFP{} for linear classification.
%\item We empirically show on vision tasks that \DNFP{} achieves state-of-the-art near-perfect AUC-ROC scores on detecting and separating unseen test data and adversarial attacks in the grey- and black-box threat models.
%
%\item We empirically show that \DNFP{} can be leveraged to achieve state-of-the-art performance on separating unseen image data (in-distribution) from adversarial examples. 
%\item We empirically show that the performance of \DNFP{} is robust to the choice of fingerprints and is effective for a wide range of choices of hyperparameters.\footnote{Code to reproduce experiments: {\tiny\url{https://www.dropbox.com/sh/jaf9rbfvr8lf9tt/AADFqfxEY3KSZyc0PNbBfATIa?dl=0}}.}
%\item Finally, we find that \DNFP{} is robust to a full-range previously known white-box adversarial attacks; we subsequently developed a new adaptive white-box attack to break \DNFP{} (See Appendix \ref{sec:expadapt}).
%\textbf{}
%\end{itemize}

%\vspace{-5pt}
\section{Related Work}
\label{sec:related}

\textbf{Detection versus Robust Prediction}
There are two broad classes of approaches to safe-guarding against adversarial data: robust prediction and detection.
A major bottleneck with robust predictions can be the inherently larger sample complexity associated with robust learning \cite{ludwig}.
Further, in \cite{tsipras2018robustness}, the authors argue that true robustness leads to depreciation in accuracy.
%
%Safe guarding against outliers largely involves distinguishing them from the training data-distribution and flagging them as anomalous (e.g. \citep{ODIN}). 

%EXTRASPACE
\textbf{Robust Prediction of Adversarial Examples} 
Several defenses based on robust learning have been proposed \cite{Aditi,Kolter, Madry}. 
While these defenses provide robustness in the threat models ($l_\infty$ bounded attacks) under which training is performed, the robustness does not carry over to other threat-models (bounded $l_0, l_1$ distortions \citep{sharma2018attacking}, attacks with generative models \citep{GAN_adv}).

%EXTRASPACE
Several other defenses attempt to make robust predictions: by relying on randomization \citep{xie2018mitigating}, introducing non-linearity that is
not differentiable \citep{buckman2018thermometer} and by relying on Generative Adversarial Networks \citep{GAN1, GAN2} for denoising images.
%EXTRASPACE
However, recent work has shown that several of these defenses are not secure \citep{obfuscated-gradients, obscurityicml, CarliniBreak}.
Several defenses proposed in the grey-box setting  \citep{magnet, cisse} have been rendered vulnerable by attacking proxy defenses \citep{CarliniMagnet}, ensembles and attacking the expectation over the distribution (for defenses relying on randomization)
\citep{obfuscated-gradients}.
We study \DNFP{} extensively under these attacks and find that \DNFP{} is robust to such adaptive grey-box attacks.

%EXTRASPACE
%EXTRASPACE
\textbf{Robust Detection of Adversarial Examples}
Amongst defenses that study the detection of adversarial examples, \cite{ma2018characterizing} detects adversarial samples using an auxiliary classifier trained to use an expansion-based measure, \emph{local intrinsic dimensionality (LID)}.
%EXTRASPACE
Similar detection methods based on Kernel Density (KD), Bayesian-Uncertainty (BU) \citep{KD} and the Mahalanobis Distance (MD) \citep{Kimin} using artifacts from trained networks have been considered.
In contrast to such methods (including \citep{magnet}), \DNFP{} does not need auxiliary classifiers and performs significantly better than LID, KD and BU when the attack mechanism is unknown.

%\textbf{Outlier Detection in Image Data} 
%In \citet{MSP}, the maximum softmax probability (MSP) is introduced as a metric to distinguish between out-of-distribution samples and samples from the training-distribution. 
%\citet{ODIN} (ODIN) improves on this by using temperature scaling and input processing with noise.
%\citet{Kimin} (Mahalanobis) show that the  Mahalanobis distance can be used as a metric to detect outliers more reliably than \citep{MSP, ODIN}.
%Further, \citet{Kimin} show that by validating the detector against adversarial examples, they are able to detect outliers reliable.
%In \citet{OE}, the authors illustrate that exposure to outliers from a different distribution than the one being detected assists in detecting outliers.
%With \DNFP{}, we do not rely on validation/exposure against any form of outliers or input pre-processing.
%The encoding of the fingerprints forces the MSP to peak on clean data, making them much more distinguishable.
%Leveraging this, for networks trained on CIFAR-100, we are able to improve detection from 44.9\% to 94.5\% on  SVHN, 56.6\% to 85.2\% on LSUN, 70.3 \% to 84.6\% on TinyImagenet.

\textbf{Adversarial Attacks in Other Domains.}
We evaluate \DNFP{} on computer vision tasks, a domain in which DNNs have proved to be effective and adversarial examples have been extensively studied.
\DNFP{} could potentially be employed to secure DNNs against attacks in domains such as speech recognition \citep{audioadv} and text comprehension \citep{jia2017adversarial}.
Adversarial attacks have also been studied in domains such as detection of malware \citep{Grosse}, spam \citep{SpamBayes} and intrusions \citep{MimicryIntrusion}.
Data-poisoning \citep{DataPoisoning} is another form of attack where maliciously crafted data is injected during training.
%\vspace{-5pt}

\begin{figure}[t]
	\includegraphics[width= 0.55\linewidth]{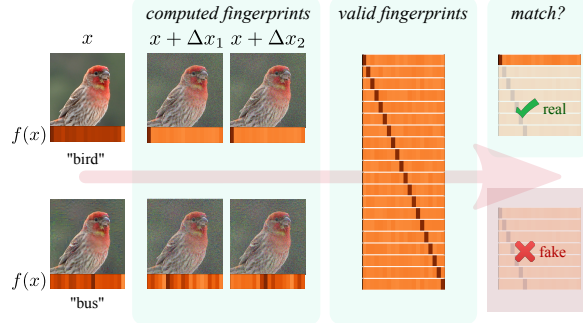}\vspace{-0.1in}
	\caption{Example of detecting adversarial examples using \DNFP{} with $N=2$ fingerprints, for $K$-class classification. 
		$\varphi(x)$ is the model output.
		\DNFP{} separates real data $x$ (top) from adversarial data $x' = x + \pert{}$ (bottom) by comparing the sensitivity of the model to predefined perturbations around unseen inputs with a reference sensitivity encoded around the manifold of \emph{real} images during training.
		The training for \DNFP{} forces the maximum softmax scores on in-distribution samples to be high, and a threshold on the maximum softmax score distinguishes between real examples from adversarial.
	}\label{fig:fp}
\end{figure}

\section{Neural Fingerprinting}

We consider supervised classification, where we aim to learn a model $f(x;\theta)$ from \emph{real} data $\brckcur{(x^i,y^{*i})}_{i\in\calI}$, where $x\in\mathbb{R}^l$ is an input example (e.g., an image) and $y^*$ is a 1-hot label vector $y^* \in \brckcur{0,1}^K$ over $K$ classes.
Here, we assume the data is sampled from a data distribution $P_{data}(x,y)$.
%EXTRASPACE
For example, a neural network $f$ predicts class probabilities $P(y|x;\theta)$ as:
%EXTRASPACE
\eq{
	%EXTRASPACE
	f(x;\theta)_j = P(y_j|x;\theta) = \fr{\exp{h(x;\theta)}_j}{\sum_l\exp{h(x;\theta)_l}},
	\label{eq:model}
}
%EXTRASPACE
where $h(x;\theta) \in \mathbb{R}^K$ are called logits and the most likely class is chosen. 
The optimal $\theta^*$ can be learned by minimizing a loss function $L(x, y; \theta)$, e.g., cross-entropy loss.
%EXTRASPACE
% The real data are sampled from the \emph{data distribution} $P_{data}(x,y)$. Therefore, we consider a sample $(x, y^{*})$ to be real for which $P_{data}(x,y^{*}) > 0$. 
%
% In this spirit, we define \emph{fake} data to be inputs that are not from the underlying data-distribution from which the training data was sampled from, i.e. inputs where the model prediction cannot be relied on. This includes outliers and adversarially crafted inputs. 
%
%\subsection{Adversarial Data}

\paragraph{Adversarial Attacks}
In an adversarial setting, an attacker attempts to construct adversarial examples $x'$, such that $\hat{y} = \argmax_l P(y_l|x';\theta)$ is an \textit{incorrect} class prediction (i.e., $P_{data}(x', \hat{y})=0$). 
%\CORR{Formally, this can be defined as $P_{data}(x', f(x';\theta)) = 0$, i.e., an adversarial example is an out-lier.}
% 
In this work, we focus on \emph{bounded} adversarial attacks, which produce small perturbations $\eta$ that cause mis-classification. This is a standard threat model, for an extensive review see \citep{adversarialsurvey}.
More generally, a bounded adversarial example causes a large change in model output, i.e. for $\delta, \rho > 0$,
%EXTRASPACE
$\norm{\eta} \leq \delta$,  we have $\norm{f(x+\eta) - f(x)} > \rho$
such that the class predicted by the model changes:
%EXTRASPACE
$
\argmax_j f(x+\eta)_j \neq \argmax_j f(x)_j.
$
%EXTRASPACE
An example bounded attack is the Fast-Gradient Sign Method (FGSM) \cite{GoodfellowSS14} which uses an input-space gradient:
$\eta \propto \sign{\fr{\partial L(x, y; \theta) }{\partial x}}.$
Since the perturbation $\eta$ is small, a perturbed image $x'$ can be indistinguishable from $x$ but still cause very different predictions. 

%
%We note that, disregarding the computational issues of an attacker, detecting adversarial attacks is harder than detecting outliers because of the potentially adaptive nature of adversarial attacks. Our goal in this work is to efficiently detect and separate real from adversarial data.

\paragraph{Encoding Fingerprints into Network Response}
To defend DNNs against adversarial data, we propose training networks with \DNFP{}. 
Training with \DNFP{} entails encoding fingerprints around the data-distribution into the network response, and the network response can then be leveraged to detect whether an input example $x$ is real or adversarial. 
A defender using \DNFP{}  begins by choosing input perturbation(s) $\Delta x$ around $x$, and desired 
changes in the model response $\Delta y$. 
Formally, we define a fingerprint $\fp{}$ as the tuple $
\fp{} \triangleq (\Delta x, \Delta y).$
%EXTRASPACE
For $K$-class classification, we define a set of $N$ fingerprints:
%EXTRASPACE
\eq{
	%EXTRASPACE
	\fp{}^{i,j} = (\fpx^{i}, \fpy^{i,j}), \hspace{5pt} i = 1,\ldots,N, \hspace{5pt} j = 1,\ldots,K,
	%EXTRASPACE
	\label{eq:fpset}
}
where $\fp{}^{i,j}$ is the $i^\textrm{th}$ fingerprint for class $j$.
%EXTRASPACE
%
To characterize sensitivity, we define the function $F(x,{\Delta x}^i)$ to measure the change in model output. A simple choice could be $F(x,\Delta x^i)=f(x+\Delta x^i)-f(x)$ (although we will use variations hereafter).
Once a defender has constructed a set of desired fingerprints $\fp{}$, the chosen fingerprints can be embedded into the network's response by adding a fingerprint regression loss during training. %EXTRASPACE
Given a classification model, the fingerprint loss is:
%EXTRASPACE
\eq{
	%EXTRASPACE
	\fploss{}(x, y, \fp{}; \theta) &= \sum_{i=1}^{N}
	%EXTRASPACE
	\norm{ F(x, \fpx^i) - \fpy^{i,k} }_2^2,
	\label{eq:encode}
}
%EXTRASPACE
where $k$ is the ground truth class for example $x$ and $\fpy^{i,k}$ are the fingerprint outputs. Note that we \emph{only train on the fingerprints for the ground truth class}.
%EXTRASPACE
The total training objective then is:
%EXTRASPACE
\eq{
	\min_{\theta} & \sum_{(x,y)} \brck{ 
		L_0(x,y;\theta) + \alpha \fploss{}(x,y,\fp{};\theta)
	},
	%EXTRASPACE
	\label{eq:fpcost}
}
%EXTRASPACE
where $L_0$ is a loss function (e.g. cross-entropy loss for classification) and $\alpha$ a positive scalar. 
Hereafter, we will use $\alpha = 1$ for brevity, but 
in practice, we choose $\alpha$ to balance the task and fingerprint losses.
%EXTRASPACE
As noted before, the $\fpx^i$ ($\fpy{}^{i,j}$) are chosen by the defender. Note that we use the same directions $\fpx^i$ for each class $j=1,2\ldots,K$, and that $\fpy{}^{i,j}$ can be either discrete or continuous depending on $f(x;\theta)$.
Here, the goal of a defender is to minimize $D(x, f, \fp{}^{i,j})$ for real data $x$. 

\paragraph{Detecting Adversarial Examples}
We exploit the characterization described above to detect adversarial examples by comparing $F(x,\Delta x^i)$ with the reference output-perturbation $\Delta y^i$:
\eq{
	D(x, f, \fp{}^{\cdot,j}) &\triangleq \fr{1}{N}\sum_{i=1}^{N} \norm{ F(x, \fpx^i) - \fpy^{i,j} }_2 \label{eq:metric1}. \ \ \ \ \  \mbox{(comparison function)}
}

\begin{wrapfigure}{r}{0.4\textwidth}
	\vspace{-0.15in}
	\begin{algorithm}[H] 
		\caption{NeuralFP}
		\label{alg:nfp}
		%\begin{small}
		\begin{algorithmic}[1]
			\STATE  {\bfseries inputs:} example $x$, model $f$, comparison func.~$D$ (Eq.~\ref{eq:metric1}), threshold $\tau>0$, fingerprints $\{\fp{}^{i,j}\}$ (Eq.~\ref{eq:fpset})
			\IF{$\exists j:\ D(x,f,\fp{}^{\cdot,j}) \leq \tau$}
			\STATE \textbf{return:} accept \ \ \ \ \greencomment{real}
			\ENDIF
			\STATE \textbf{return:} reject\ \ \ \ \redcomment{fake}
		\end{algorithmic}
		%	\end{small}
		\vspace{-5pt}
	\end{algorithm}
	\vspace{-5pt}
\end{wrapfigure}
The objective from  \eqref{eq:fpcost} trains the model so that the function $D$ has low values around the real data, e.g., the train-set.
The approach for detecting adversarial examples is summarized in Algorithm \ref{alg:nfp} and Figure \ref{fig:fp}. 
The key idea is to detect adversarial examples by using a \emph{consistency check of the model output around the input}. 
\DNFP{} classifies a new input $x'$ as \textit{real} if the change in model output  is close to the $\fpy^{i,j}$ \emph{for some class $j$}, for all $i$.
Here, we use $D$ and threshold $\tau>0$ to define the level of agreement required, i.e., we declare $x'$ \textit{real} when $D$ is below a threshold $\tau$:
\smallskip
%EXTRASPACE
\eq{
	x' \textrm{ is real} \Leftrightarrow \exists j: D(x', f, \fp{}^{\cdot,j}) \leq \tau.
	%EXTRASPACE
	\label{eq:cond}
}
%EXTRASPACE
Hence, the \DNFP{} test is defined by: $\FPdata{} = \brck{\brckcur{\fpxy{}^{i,j}}_{i=1\ldots N, j=1 \ldots K}, D, \tau}$.
% 
%
%\paragraph{Detecting Outliers} We choose the $\Delta y$ such that the maximum softmax score peaks for real-data. This allows us to utilize the MSP \citep{MSP} to distinguish between outliers and real-data.
%The exact choice of fingerprints is discussed in detail in Section \ref{sec:fp_choice}.
% Formally, for a threshold $\tau$, a data-point is classified as real if and only if: 
%\[
%\underset{j}\max \qquad %\fr{\exp{h(x;\theta)}_j}{\sum_l\exp{h(x;\theta)_l}} \geq %\tau.
%\]

\paragraph{Computational Complexity}
%EXTRASPACE
Extra computation comes from checking the condition in \eqref{eq:cond} which requires $O(NK)$ forward passes to compute $F(x, \Delta x^i)$. A straightforward implementation is to check \eqref{eq:cond} iteratively, and stop whenever an agreement is seen or all classes have been exhausted. One can also parallelize the computation  and use minibatches for real-time applications.

\subsection{Choosing and Characterizing Fingerprints: Linear Models}
\label{sec:choosecharfp}
%EXTRASPACE
% We first theoretically analyze \DNFP{} for linear models and how fingerprints characterize data distributions. In particular, given \FPdata{} as in \refn{nfpdata}, which inputs $x$ will be classified as ``real'', and how  $N$ of fingerprints and how efficient fingerprints are?
%EXTRASPACE
% \paragraph{Linear Models.}
%EXTRASPACE
\begin{wrapfigure}{r}{0.36\linewidth}
	\vspace{-0.2in}
	\includegraphics[width=\linewidth]{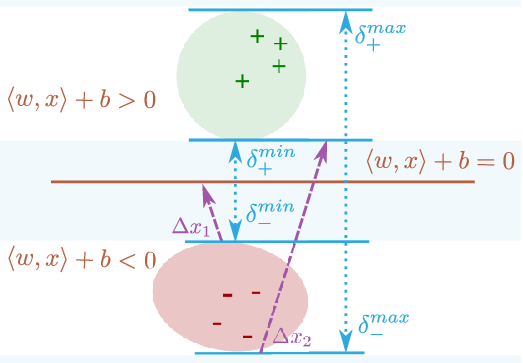}
	\vspace{-15pt}
	\caption{Geometry of fingerprints for SVMs with linearly separable data. Let $d(x)$ be the signed distance to the decision boundary (see Thm \ref{thm:svm}), and $\delta^{max}_{\pm}$ ($\delta^{min}_{\pm}$) denote the maximal (minimal) distances of the positive ($x_+$) and negative ($x_-$) examples to the hyperplane $\brcka{w,x} + b = 0$. The fingerprint $\fpx^{1}$ with $\brcka{\fpx^{1},\hat{w}} = \delta^{min}_-$ will have $f(x_- + \fpx) < 0$ and $f(x_-) < 0$ for all $x_-$ in the data distribution (red region). Hence, $\fpx^{1}$ will flag all $x'$ in the regions $-\delta^{min}_- < d(x') < 0$ as not real, since for those $x'$ it will \emph{always} see a change in predicted class. Similarly, $\fpx^2$ with $\brcka{\fpx^2,\hat{w}}=\delta^{max}_-$ always sees a class change for real $x_-$, flagging all $x': d(x') < -\delta^{max}_-$ as not real.
		\vspace{-30pt}
	}
	\label{fig:svm}
\end{wrapfigure}
%EXTRASPACE
We first analyze \DNFP{} on binary classification with data $\brckcur{\brck{x^i, y^{*i}}}_{i\in\calI}$ and linear model (SVM):
%EXTRASPACE
\eq{
	\nonumber
	f(x) = \brcka{w, x} + b, \hspace{10pt} \hat{y} = \sign{f(x) \in \brckcur{-1,1},}
}
%EXTRASPACE
on inputs $x^i\in\mathbb{R}^n$, where $n \gg 1$ (e.g., $n=900$ for MNIST).
%EXTRASPACE
The SVM defines a hyperplane $f(x) = 0$ to separate positive ($\hat{y} = +1$) from negative examples ($\hat{y} = -1$).
%EXTRASPACE
We will assume that the positive and negative examples are linearly separable by a hyperplane defined by a normal $\hat{w} = \fr{w}{\norm{w}}$.
%EXTRASPACE
We define the minimal and maximal distance from the examples to the hyperplane along $\hat{w}$ as:
%EXTRASPACE
\begin{flalign}
%EXTRASPACE
\nonumber
\delta_{\pm}^{min} = \min_{i:y^i = \pm 1} \abs{\brcka{x^i, \hat{w}}}, \hspace{2pt} \delta_{\pm}^{max} = \max_{i:y^i = \pm 1} \abs{\brcka{x^i, \hat{w}}}.
%EXTRASPACE
\end{flalign}
%EXTRASPACE
In this setting, the set of $x^i$ classified as real by fingerprints is determined by the geometry of $f(x)$. 
Here, for detection, we measure the \emph{exact change in predicted class} using: 
\begin{align*}
F(x,{\Delta x})= \Delta y = &\ \textrm{sign} \brck{\brcka{w,x+\Delta x}+b}\\
&\ - \textrm{sign} \brck{\brcka{w,x}+b} \in \brckcur{-2,0,2}.
\end{align*}
%EXTRASPACE
\begin{thm}[Fingerprints for SVM]
	\label{thm:svm}
	Consider an SVM with $\hat{w} = \fr{w}{\norm{w}}$ and separable data, and the following:
	\begin{align}
	(\Delta x^1 &= \delta^{min}_- \hat{w}, \Delta y^{1,-} = 0), \label{fp3} \\ 
	(\Delta x^2 &= \delta^{max}_- \hat{w}, \Delta y^{2,-} = +2), \label{fp4} \\
	(\Delta x^3 &= -\delta^{max}_+ \hat{w}, \Delta y^{3,+} = -2), \label{fp1} \\
	(\Delta x^4 &= -\delta^{min}_+ \hat{w}, \Delta y^{4,+} = 0). \label{fp2}
	\end{align}
	An adversarial input $x'=x_{\pm} + \pert{}$ is one for which one of the following holds:
	%EXTRASPACE
	\begin{eqnarray}
	d(x') > \delta^{max}_+,\hspace{10pt}
	0 < d(x') < \delta^{min}_+,\hspace{10pt}
	d(x') < -\delta^{max}_-,\hspace{10pt}
	-\delta^{min}_- < d(x') < 0.
	\label{svmregions}
	\end{eqnarray}
	%will satisfy one of the above criteria. 
	Here, $d(x') = \frac{ \brcka{x', w} + b}{\norm{w}}$ is the signed distance to the separating hyperplane. 
\end{thm}
The proof for two fingerprints is shown in Figure \ref{fig:svm}. The full proof is in the Appendix.
%EXTRASPACE
Theorem \ref{thm:svm} by itself \textit{does not} detect adversarial inputs that involve  distortions \emph{parallel} to the decision boundary. An adversary could push a negative example $x_-$ across the boundary to a region outside the data distribution ($P_{data}(x_- + \eta,y)=0$), but within distances $\delta_+^{min}$ and $\delta_+^{max}$ of the boundary. This would still be judged as real by using fingerprints. However, such examples could still be detected by also checking the distance of $x_- + \eta$ to the nearest $x_+$ in the dataset.
%EXTRASPACE

% \paragraph{Non-linear Models}

% 

\subsection{Choosing and Characterizing Fingerprints: DNNs}
\label{sec:fp_choice}
% Theorem \ref{thm:svm} assumed that the data was perfectly separable. 
%
%
% We focus on neural networks, high-capacity models that can learn complex patterns \cite{zhang2016understanding}\footnote{In fact, \cite{zhang2016understanding} showed that neural networks can fit arbitrary random or complex patterns of prediction labels.} and enforce a specific desired change in the (normalized) logits along randomly sampled directions during training (equation \eqref{eq:encode}).
%
In contrast to the linear setting, in general \DNFP{} utilizes a \emph{softer} notion of fingerprint matching by checking whether the model outputs match \emph{changes in normalized-logits}.
Specifically, for classification models $f(x;\theta)$ with logits $h(x;\theta)$ (see Eqn \ref{eq:model}), where $F$ is defined as: 
\eq{
	\nonumber
	F(x,{\Delta x}^i) &\triangleq \varphi(x+\Delta x^i) - \varphi(x),\hspace{10pt} \varphi(x) \triangleq h(x;\theta)/\norm{ h(x;\theta)},
}
where $\varphi$ are the normalized logits. 
The logits are normalized so the DNN does not fit the $\Delta y$ by making the weights arbitrarily large.
Here, we use $D(x,\Delta x^i)$ as in \refn{eq:metric1}. Note that here $\fpy{}^{i,j} \in \mathbb{R}^K$.

\textbf{Choosing $\Delta y$}
For our experiments, we choose the $\Delta y$ so that the normalized-logit of the true class either increases or decreases along the $\Delta x^i$ (analogous to the linear case). For e.g., for a 10-class classification task, if $x$ is in class $k$ we choose $\Delta y$ of the form:
%\eq{\label{eq:nonlinearfp}
$\Delta y^k_{l\not=k} = \alpha$ and $\Delta y^k_{l=k} = -\beta$, for $k = 1, \ldots, 10$, and
%}
$\alpha, \beta \in \mathbb{R}^+$.
We found that reasonable choices are $\alpha=0.25$ and $\beta=0.75$, and that the method is not sensitive to these choices.
For investigation in the grey-box setting (with an adaptive attacker), we randomize the $\Delta y^k$ in addition to the $\Delta x$.
This randomization minimizes the assumptions the attacker can make about the fingerprints and is detailed in Appendix \ref{supp:sec:greyboxparams}.

% 
% This is consistent with the observation in \cite{zhang2016understanding} that DNNs can fit arbitrary input-output data.
% 
\textbf{Choosing $\Delta x$}
For nonlinear models (e.g., DNNs), the best fingerprint-direction $\Delta x$ choice is not clear.
We use a straightforward extension from the linear case with randomly sampled $\Delta x$'s. Randomization minimizes structural assumptions that may make \DNFP{} exploitable. %, and we find empirically 
For all experiments, we sampled $\Delta x^i$ uniformly  from $[-\varepsilon,\varepsilon]$: $\fpx^i \sim \mathcal{U}(-\varepsilon,\varepsilon)^l$, where $l$ is the input dimension. 
Our experiments (see Figure \ref{f:multiruns}) suggest that \DNFP{} is \emph{not sensitive} to the random values sampled.
This also suggests that the $\Delta x^i$ could be chosen based on different approaches.
% This suggests that \DNFP{} is effective with (randomly sampled) fingerprints as well.}
%
% The high model-capacity of neural networks allows to learn such complex patterns, that can be hard to reverse engineer. 
%
%EXTRASPACE
%\paragraph{Nonlinear Models.}
%{
%\Sumanth{\begin{itemize}
%\item Fingerprinting creates valleys of lower fingerprint loss around the data-distribution.
%\end{itemize}}}
%{\Sumanth{Add experimental visualizations and remove text maybe}}

%For the case of non-linear models (e.g. DNNs) it is difficult to identify the fingerprint-directions from the model itself, as in the linear case.
%
%For this setting, we use highly expressive models and train them to have valleys of low fingerprint-loss around the \emph{real} data distribution.
%
%By checking if inputs belong to a region with a low fingeprint-loss, we are able to ascertain if the input belongs to \emph{real} data manifold.

%For the general setting, e.g. for nonlinear models $f$, Theorem \ref{thm:svm} can be extended if the data is (locally) separable in some feature space $\Phi$ and we can write a general (local) model as
%EXTRASPACE
%\eq{y = \brcka{w,\phi(x)} + b.}
%EXTRASPACE
%In this case, the fingerprints can be defined analogous to (\ref{fp1}-\ref{fp4}). As such, it is straightforward to lift the %analysis of Theorem \ref{thm:svm} to this setting as well. However, depending on the feature space $\Phi$ chosen, such as %analysis might only be applicable to a \emph{local} region of the input space and require more complex fingerprints.

%\begin{wrapfigure}{r}{0.6\linewidth}
\begin{figure}[t]
	%\vspace{-0.1in}
	\centering 
	\includegraphics[width=.245\linewidth,height=0.19\linewidth]{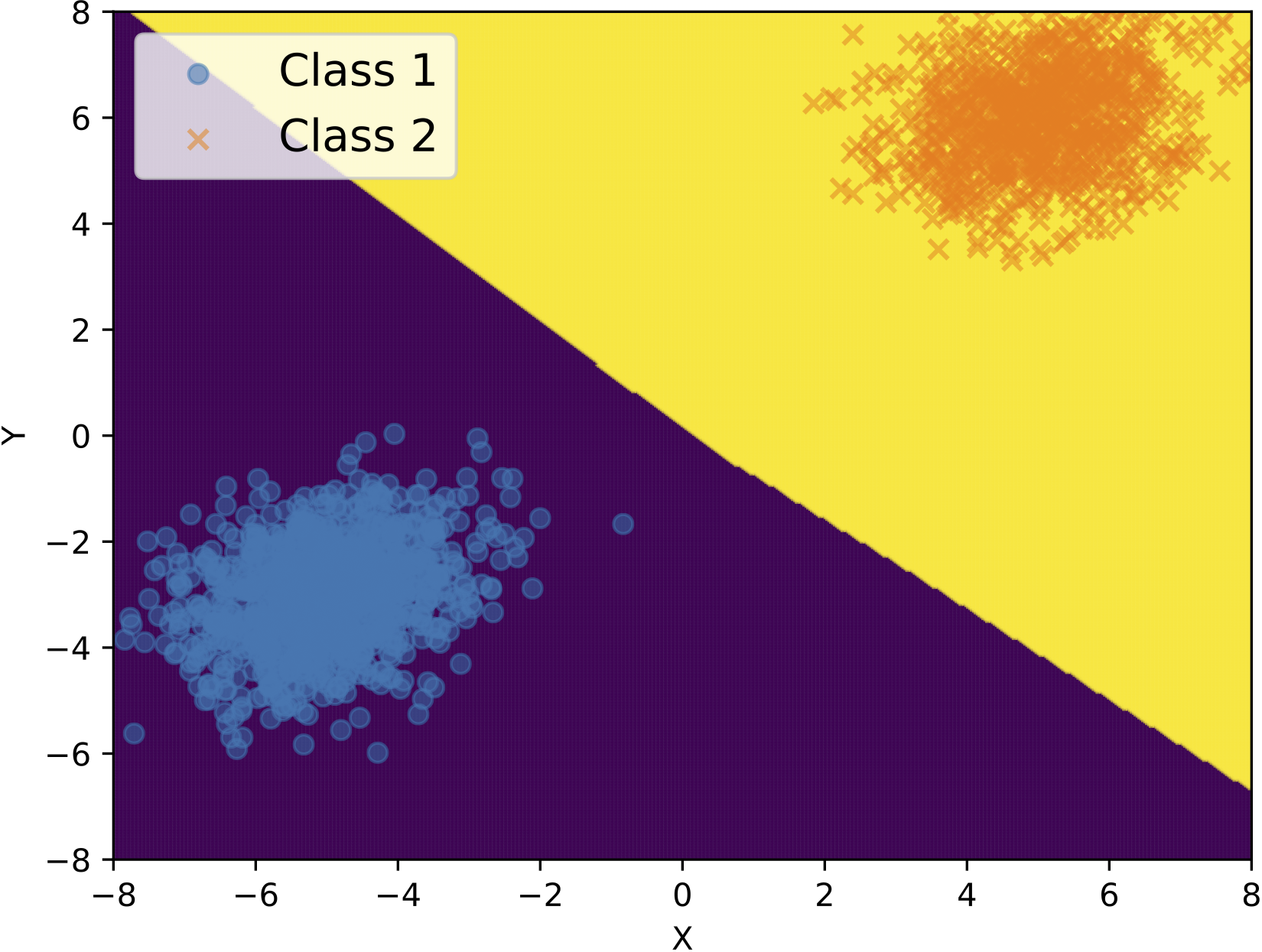}
	\includegraphics[width=.245\linewidth,height=0.19\linewidth]{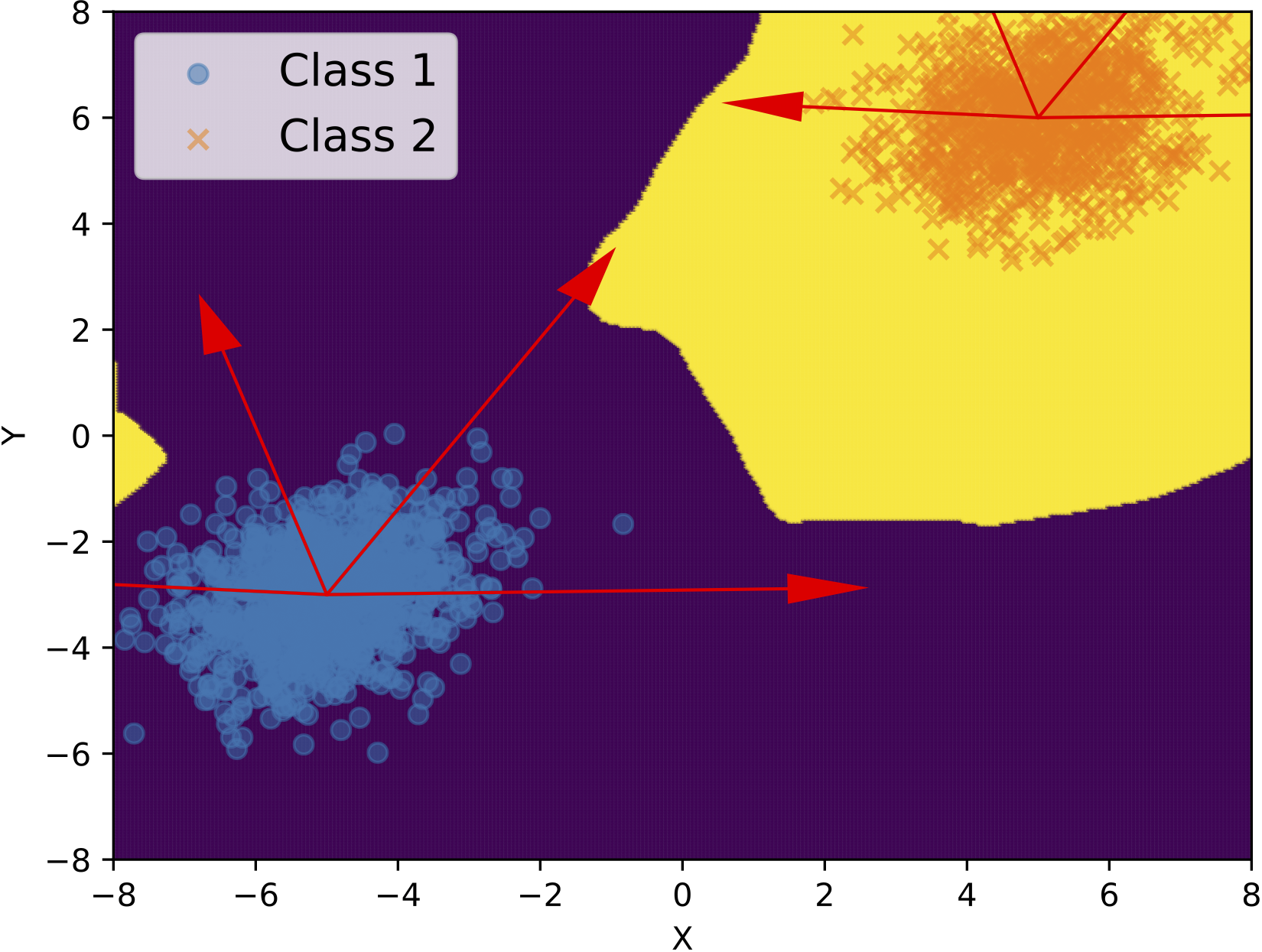}
	\includegraphics[width=.245\linewidth,height=0.19\linewidth]{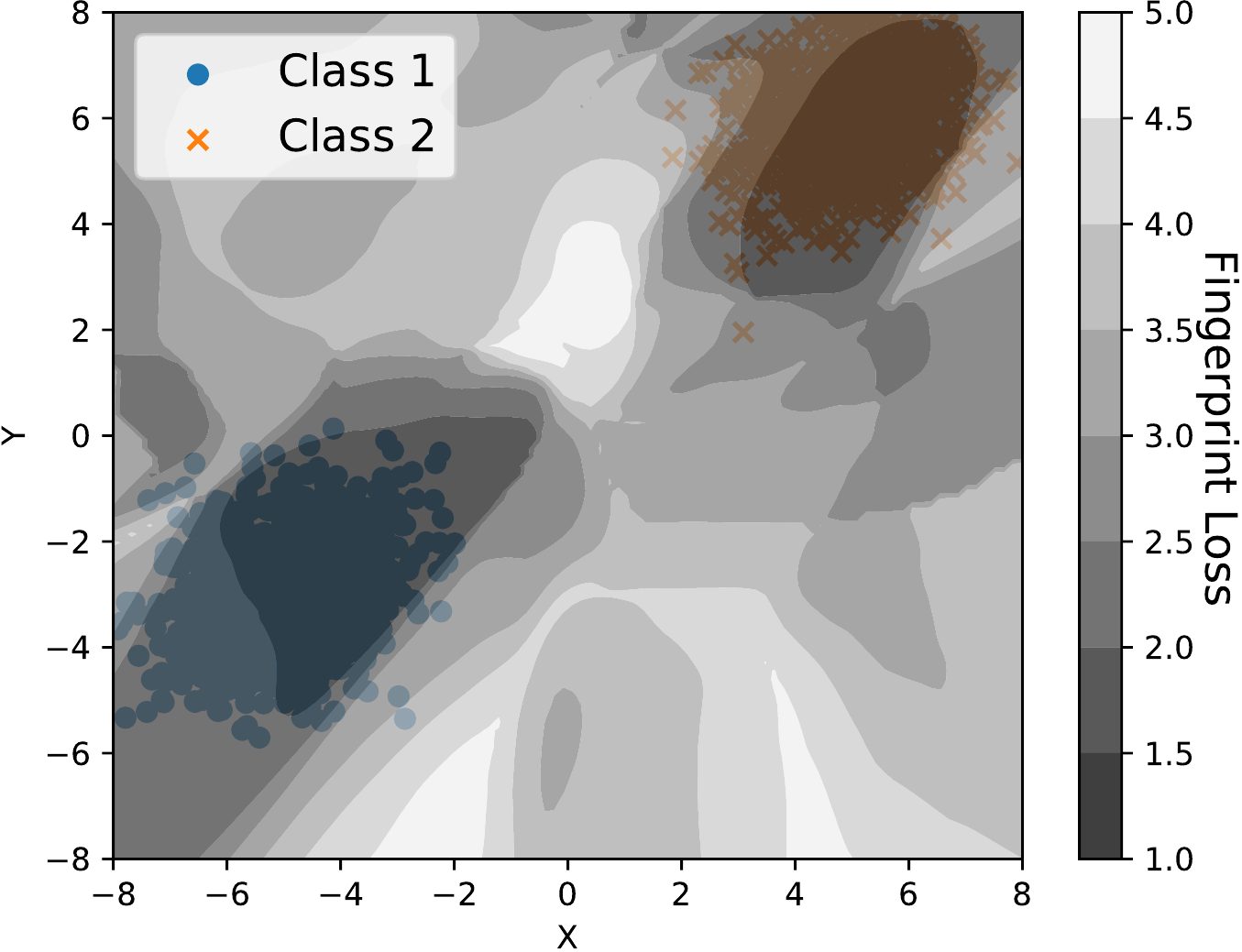}
	{\includegraphics[width=0.245\linewidth,height=0.19\linewidth]{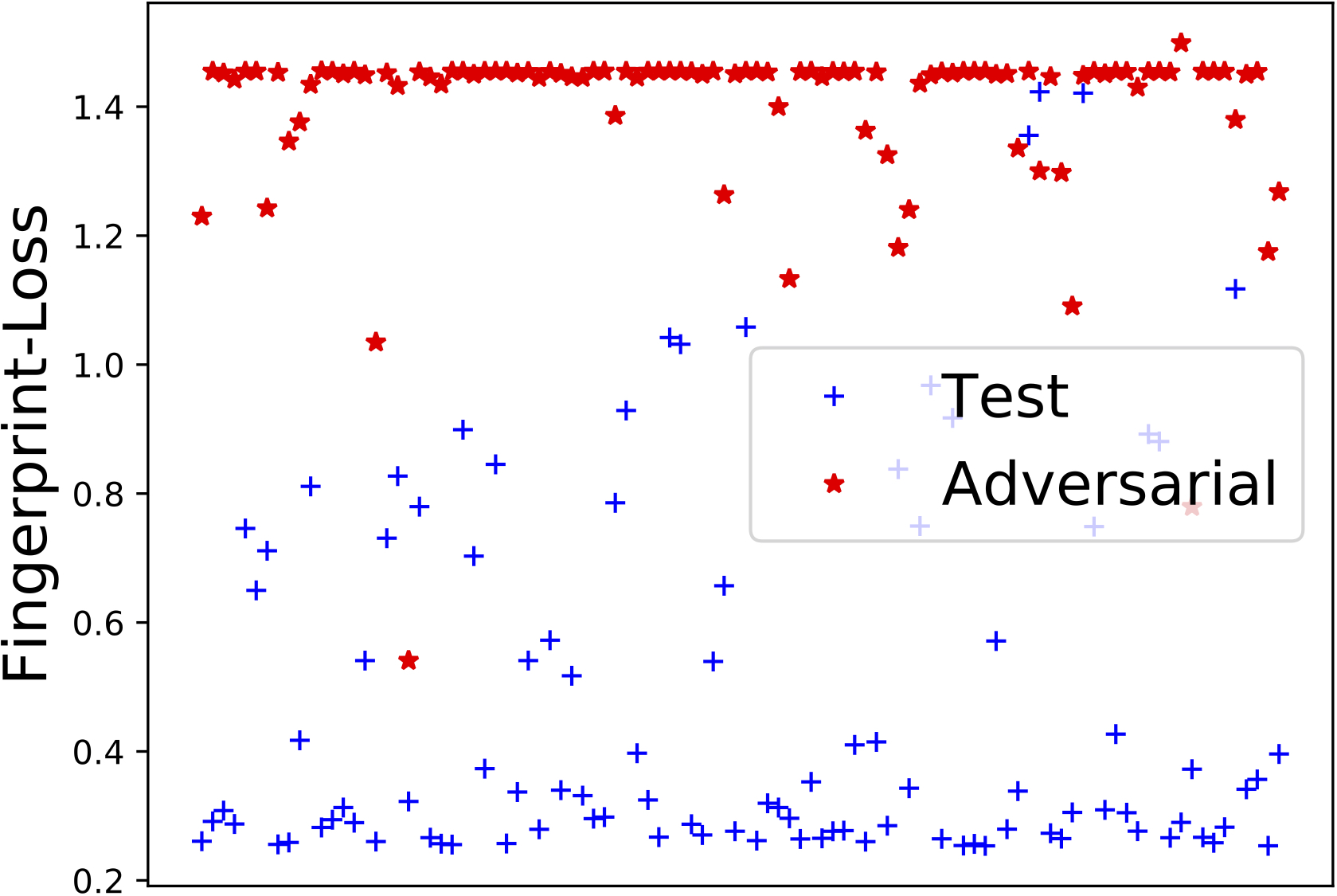}}
	\vspace{-0.2in}
	\caption{
		Left to right:
		1) decision boundary without fingerprints.
		2) with fingerprints, red arrows indicate fingerprint-directions. The decision boundary is more non-linear.
		3) contour plot of fingerprint loss. \DNFP{} detects dark regions as ``real'', while lighter ones are ``fake'' (tunable through $\tau$). Fingerprinting creates valleys of low-loss delineating real data.
		4) We see a similar phenomenon for the CIFAR-10 dataset. Fingerprint losses on 100 random test (blue) and adversarial (red) images (Table \ref{table:auc_roc}). We see a clear separation, illustrating that \DNFP{} is effective across many thresholds $\tau$.
		% \vspace{-15pt}
	}
	\label{fig:visualizefp}
	%\end{wrapfigure}
\end{figure}
\textbf{Visualizing Fingerprints.}
To understand the behavior for non-linear models, we trained a DNN (two hidden layers with 200 ReLU nodes each) to distinguish between two Gaussian balls in 2D spac.  Figure \ref{fig:visualizefp} describes this experiment.
Without fingerprints, the model learns an almost linear boundary separating the two balls (compare with Figure \ref{fig:svm}). 
When we train to encode the fingerprints, 
%(negative of $\Delta y$'s in \refn{eq:nonlinearfp}), 
we observe that \DNFP{} causes the model to learn a non-linear pockets of low fingerprint-loss characterizing the data-distribution.
In this simple setting, \DNFP{} learns to delineate the data-distribution, where the darker regions are accepted as real and the rest is rejected.

\subsection{Threat Model Analysis for Adversarial Attacks}
\label{sec:threatmodel}
% 
%\begin{wraptable}{r}{.3\linewidth}
%\vspace{-20pt}
%\end{wraptable}
% \vspace{-5pt}
% 
%EXTRASPACE
%\yy{This feels out of place, given that we're also considering outliers.  Maybe we can make explicit that this subsection is focused on the adversarial setting?}

In the adversarial setting, the various threat-models describe varying amounts of information available to the attacker.
We study \DNFP{} under multiple such threat models (Table \ref{tab:threatmodel}),
where the attacker has varying levels of knowledge of $\FPdata{}$ and model $f(x; \theta)$.

In the \textbf{black-box attack setting} the attacker has no knowledge of the network weights\ \FPdata{}. 
We study \DNFP{} in this setting, and discuss query and transfer based attacks. 

\begin{wraptable}{r}{0.4\linewidth}
	\vspace{-0.05in}
	\begin{tabular}{ccl}
		$\theta$ & \FPdata{} &   \\
		\midrule 
		$\times$ & $\times$ & black-box attack \\
		\checkmark &  $\times$ & grey-box attack \\
		\checkmark & \checkmark  & white-box attack
	\end{tabular}
	\vspace{-0.1in}
	\caption{Threat-models: attacker knows $\theta$ and/or \FPdata{}. \vspace{-5pt}
	}
	\label{tab:threatmodel}
\end{wraptable}
In the \textbf{grey-box} setting, the attacker has access to $\theta$, can query $f(x;\theta)$ and its derivatives, \emph{but does not know \FPdata{}}.
We assume the adversary has access to the distribution from which the fingerprints are sampled but not the exact sequence of the fingerprints.
This is a frequently used  threat-model in cryptography and in  practice, where a key or a password is used to secure a system.
%EXTRASPACE
%We evaluate this setting extensively in Section \ref{sec:exp_pwa}. 
%
This setting is the most commonly studied, and is the focus of most defenses reported in Section \ref{sec:related} (e.g., \citep{ma2018characterizing,GAN1, Kimin}).
The grey-box setting is relevant, for instance, when the attacker has a copy of the classifier, but the fingerprints are private, in accordance with Kerchhoff's principle \citep{Kerchoff} (e.g. when \DNFP{} is run on a cloud).
We study grey-box attacks in detail because of its practical relevance, and show that safe-keeping of the fingerprints ensures robustness against adversarial attacks. 

We note that reverse engineering the \FPdata{} by brute-force search can be combinatorially hard. To see this, consider a simple setting where only the $\fpy^{i.j}$ are unknown, and that the attacker knows that each fingerprint is discrete, i.e. each component $\fpy^{i.j}_k = \pm 1$.
Then the attacker would have to search over combinatorially ($\calO(2^{NK})$) many $\fpy$  to find the subset of $\fpy$ that satisfies the detection criterion in  \refn{eq:cond}. 
Further, smaller $\tau$s reduce the volume of inputs accepted as real. 
In our investigation in the grey-box setting, we randomize both the set of $\Delta x$s and the $\Delta y$s, thereby minimizing the structural assumptions the adversary can make with regard to the fingerprints.

The \emph{strongest} threat model assumes the attacker has full information about the defender, for e.g., when the attacker can either reverse-engineer or has access to the fingerprint data $\FPdata{}$, so that stronger attacks could be possible. 
In this setting, \DNFP{} is robust to the full range of attacks in literature.
However, we find that a new attack that adaptively solves the multi-objective optimization problem of minimizing the fingerprint-loss while producing a misclassification is able to fool \DNFP{} at large computational costs.
We discuss the evaluation in detail under this setting in Appendix \ref{sec:expadapt}.
%A defender can shrink this space of accepted  by setting the threshold level of $\tau$. 
%

%%
Additionally, we define the notion of \textbf{white-box defense} (defender is aware of the attack mechanism) and \textbf{black-box defense} (defender has no knowledge about attacker) -- \DNFP{} is a black-box defense. 
There has been considerable progress in black-box attack and white-box defense settings (e.g. \citep{guideddenoising}). However, progress in grey-box attack, {black-box defense} threat-models is relatively limited.

%EXTRASPACE

%\vspace{-5pt}
% \begin{figure}
% \floatbox[{\capbeside\thisfloatsetup{capbesideposition={left,top},capbesidewidth=6cm}}]{figure}[\FBwidth]
% {\caption{AUC-ROC mean $\mu$ and standard-deviation $\sigma$ for 32 randomly sampled fingerprints (including randomizing $N$) for CIFAR-10. The AUC-ROC across all attacks varies little ($\sigma<1\%$), with $\sigma$ highest for CW-${L_2}$, suggesting $\DNFP{}$ is robust across hyperparameters.
% }\label{fig:svm}}
% {\includegraphics[width=7cm]{images/auc_roc_runs_err_bar.pdf}}
% \end{figure}
% 
% 
% 
% 
%EXTRASPACE
%\section{Evaluating \DNFP{} on detection of anomalous data}
\section{Experiments}

\begin{table}[t]
	\caption{\DNFP{} detection AUC-ROC against \emph{black-box attackers} (know dataset but not model or fingerprints), on MNIST, CIFAR-10 tasks on test-set (``real'') and corresponding black-box adversarial (``fake'') samples (1328 \emph{pre-test} samples each).
		\DNFP{} achieves near perfect AUC-ROC.
		For CIFAR-10, the hyperparameters are $(N, \varepsilon)==(30, 0.003)$ . For MNIST,  $(N,\varepsilon)=(10,0.03)$.
		%
		%\CORR{Note that whitebox LID detection is unfair comparison, because \DNFP{} is blackbox-defense.}
	}
	\label{table:blackboxattack}
	\centering
	\begin{small}
		\begin{tabular}{llllllc}
			\multirow{2}{*}{\backslashbox{Dataset}{Source}}
			& Natural  &  Advers. & \DNFP{}   & Natural  &  Advers. & \DNFP{} \\
			& (FGSM) & (FGSM) & (FGSM) & (BIM-b) & (BIM-b) & (BIM-b)\\
			\hline
			{MNIST}
			& {99.96} & {99.91} & {99.32} & {99.62} & {99.77} & {99.83} \\
			\hline
			{CIFAR-10}
			& {99.92}  &  {99.89} & {99.43} & {99.71} & {99.41} & {99.35} \\
		\end{tabular}
	\end{small}
	\vspace{-0.1in}
\end{table}

%However, LID only outperforms \DNFP{} slightly in the whitebox setting, i.e. with data augmentation with CW-${L_2}$ adversarial examples, while \DNFP{} is always attack agnostic. On MiniImagenet-20, \DNFP{} achieves near-perfect AUC-ROC versus FGM and BIM-b. Note, for MiniImagenet-20, 1) attacks are prohibitively slow (JSMA, CW-$L_2$) and 2) due to time constraints, LID scores were not obtainable.
%EXTRASPACE
% 

% 

%  
\begin{table}[t]
	\caption{Detection AUC-ROC of black-box defenders (do not know attack strategy) against \emph{partial-white-box-attackers} (know model $f(x;\theta)$, but not defense details; see Section \ref{sec:threatmodel}), on MNIST, CIFAR-10 on test-set (``real'') and corresponding adversarial (``fake'') samples (1328 \emph{pre-test} samples each). \DNFP{} outperforms baselines (LID, KD, BU) on MNIST \& CIFAR-10.}
	\label{table:auc_roc}
	\centering
	\begin{small}
		\begin{tabular}{llccccc}
			Data & Method & FGSM & JSMA & BIM-a & BIM-b & CW-${L_2}$ \\
			\hline
			\multirow{2}{*}{MNIST}
			& LID  & 99.68 & 96.36 & 99.05 & 99.72 & 98.66 \\
			& KD  & 81.84 & 66.69 & 99.39 & 99.84 & 96.94 \\
			& BU  & 27.21 & 12.27 & 6.55 & 23.30 & 19.09 \\
			& KD+BU  & 82.93 & 47.33 & 95.98 & 99.82 & 85.68 \\
			%& LID & Whitebox & 99.68 & 98.67 & 99.61 & 99.90 & 99.55 \\
			& \DNFP{}  & \bf{100.0} & \bf{99.97} & \bf{99.94} & \bf{99.98} & \bf{99.74} \\
			\hline
			\multirow{2}{*}{CIFAR-10}
			& LID  & 82.38 & 89.93 & 82.51 & 91.61 & 93.32 \\
			%& LID & Whitebox& 82.38 & 95.87 & 82.30 & 99.78 & \bf{98.94} \\
			& KD  & 62.76 & 84.54 & 69.08 & 89.66 & 90.77 \\
			& BU  & 71.73 & 84.95 & 82.23 & 3.26 & 89.89 \\
			& KD+BU  & 71.40 & 84.49 & 82.07 & 1.1 & 89.30 \\
			& \DNFP{}  & \bf{99.96} & \bf{99.91} & \bf{99.91} & \bf{99.95} & \bf{98.87} \\
		\end{tabular}
	\end{small}
	\vspace{-0.1in}
	% \vspace{-5pt}
\end{table}

\label{sec:exp_pwa}
%EXTRASPACE
We  empirically validate \DNFP{} on a number of vision data-sets, We analyze the behavior and robustness of \DNFP{} against adversarial attacks under various threat models. 
We also study the sensitivity of \DNFP{} to varying hyperparameters.
The study of the white-box-attack setting is deferred to Appendix \ref{sec:expadapt}.
%
% Lastly, we study the robustness of \DNFP{} under various adaptive-whitebox-attacks, including a recent gradient-free attack \citep{obscurityicml}.
% 
We empirically find that:
\begin{itemize}
	\item \DNFP{} is robust under the black-box and grey-box adversarial threat-models.
	\item Using \DNFP{} does not diminish prediction accuracy (when not abstaining).
	% The test accuracy with \DNFP{} is $85 \pm 1\%$ on CIFAR-10 and $99.2 \pm 0.1\%$ on MNIST. This matches the accuracy for the same models when trained without fingerprints. 
\end{itemize}
%

%\subsection{Detecting Adversarial Examples}
\paragraph{Adversarial Attacks} We perform evaluation against adversarial attacks under various threat-models on the MNIST, CIFAR-10 and MiniImagenet-20 datasets. We use the following state-of-the-art attacks (Table \ref{table:attackparam}):
\begin{itemize}
	\item \emph{Fast Gradient Method} (FGSM) \cite{GoodfellowSS14} and \emph{Basic Iterative Method} (BIM) \cite{BIM} are both gradient based attacks with BIM being an iterative variant of FGSM. We consider both BIM-a (iterates until misclassification has been achieved) and BIM-b (iterates 50 times).
	%\footnote{Note that BIM is also known as PGD (projected gradient descent) or $\mathrm{FGSM}^\mathrm{k}.$}.
	\item \emph{Jacobian-based Saliency Map Attack} (JSMA) \cite{JSMA} perturbs pixels using a saliency map.
	\item \emph{Carlini-Wagner Attack} (CW-${L_2}$): an attack that optimizes to minimize the perturbation needed for misclassification, and is one of the strongest known attacks \citep{2016arXiv160804644C,CarliniBreak}.
\end{itemize}
On CIFAR-10 and MiniImagenet-20, all attacks have a successful misclassification rate of $99-100\%$
and on MNIST, FGSM has a successful misclassification rate of $89\%$, while the other attacks have near-perfect misclassification rates. 
See Table \ref{table:attackparam} for attack success rates and distortion bounds.

\subsection{Black-box adversarial Examples}
We consider black-box adversarial examples on  MNIST and CIFAR-10. Even though the black-box setting is the weakest threat model, practical black-box attacks have been shown to be feasible in several real-world applications \citep{transfer, practical}.

\textbf{MNIST.}
%EXTRASPACE
We train a 5-layer ConvNet to $99.2 \pm 0.1\%$ test-accuracy.
%EXTRASPACE
The set of $\Delta x^i \in \mathbb{R}^{28 \times 28}$ is chosen randomly, with each value chosen uniformly in $[-\varepsilon,\varepsilon]$.
%EXTRASPACE
For each $\Delta x^i$, if $x$ is of label-class $k$, $\Delta y^{i,k} \in \mathbb{R}^{10}$ is chosen to be such that $\Delta y^{i,k}_{l \not = k} = 0.25$ and $\Delta y^{i,k}_{l = k} = -0.75$, with $\|\Delta y\|_2=1$.
%EXTRASPACE
%

\textbf{CIFAR-10.}
%EXTRASPACE
For CIFAR-10, we trained a 7-layer ConvNet (similar to \cite{2016arXiv160804644C}) to $85 \pm 1 \%$ accuracy.
%EXTRASPACE
The $\fpx^i$ and $\fpy^{i,j}$ are chosen similarly as for MNIST.
First, we consider transfer based attacks \cite{transfer} from models trained i) naturally, ii) adversarially using robust learning, and iii) with \DNFP{}.
The results are summarized in Table \ref{table:blackboxattack}, and we see that \DNFP{} is able to detect transfer based attacks with near-perfect accuracy.
Next, we consider query based attacks such as \cite{nesblackbox,decisionbased}. 
A key first step for these attacks is the ability to sample adversarial points that can fool the detector. 
To evaluate the feasibility of such attacks on \DNFP{}, we sample large numbers of points randomly and check if they are adversarial.
We see that from over $10^9$ sampled points (with relatively large perturbations), we do not encounter a single adversarial point with a small fingerprint loss (See Appendix \ref{supp:sec:randomsampling} for details).

We finally consider an attack where the adversary has access to the distribution from which the fingerprints are sampled but not the exact fingerprints nor the model weights.
Under this scenario, the adversary can attack an ensemble of models trained with \DNFP{} by sampling fingerprints from the distribution and exploit the transferability of adversarial examples to attack the original model. 
We train an ensemble of three substitute models, and generate adversarial examples using the white-box-attack described in Appendix \ref{sec:expadapt}.
For this attack, $45.1\%$ of the adversarial examples do not transfer, and 
\DNFP{} achieves an AUC-ROC of $94.72\%$ on the remaining samples. 

\begin{table}[t]
	\caption{Detection AUC-ROC of \DNFP{} vs grey-box EOT attacks (100 pre-test samples). 
		\label{table:greyboxEOT}
	}
	\centering
	\begin{small}
		\begin{tabular}{lcc}
			Data & Unknown $\Delta x$, Unknown  $\Delta y$ & Unknown $\Delta x$, known $\Delta y$ \\
			\hline
			{CIFAR-10}
			& 100.00 & 99.85
		\end{tabular}
	\end{small}
\end{table}
\vspace{-7pt}

%
%\begin{figure}
%\centering
%\includegraphics[width=0.3\linewidth]{images/cdf_ood.pdf}
%\caption{Cumulative distribution of the MSP on CIFAR-100 (in-dist), SVHN (out-dist) datasets for ResNet18 trained on the CIFAR-100 task with \DNFP{}. The in- \& out-dist. maximum softmax probabilities are well separated. Most CIFAR-100 samples have MSP close to 1.0 while MSP for SVHN data is much smaller.}
%\end{figure}

%
%
%

\subsection{Grey-box adversarial Examples}

\label{sec:effective}
This threat-model is considerably stronger than the black-box setting.
The adversary has access to the model weights and the distribution from which the fingerprints are sampled.
The baselines LID \cite{ma2018characterizing}, a recent detection based defense; KD; BU \cite{KD}; all trained on FGSM, as in \cite{ma2018characterizing} evaluate primarily in this setting.
%EXTRASPACE
%
%
Following \cite{ma2018characterizing}, for each dataset we consider a randomly sampled \emph{pre-test}-set of unseen $1328$ test-set images, and discard misclassified pre-test images.
%EXTRASPACE
For the \emph{test-set} of remaining images, we generate adversarial perturbations by applying each of the above mentioned attacks.
%EXTRASPACE
We report AUC-ROC on sets composed in equal parts of the \emph{test-set} and \emph{test-set} adversarial samples. 
The AUC-ROC is computed by varying the threshold $\tau$.
%EXTRASPACE
See Appendix for model and dataset details.

% Table \ref{table:minim} compares \DNFP{} with baselines in a setting where the baselines use information about the attacker while \DNFP{} does not.
% 

%EXTRASPACE
%EXTRASPACE
%
The FGSM, BIM-a, BIM-b and JSMA attacks are untargeted. 
We use published code for the attacks and code from \cite{ma2018characterizing} for the baselines. 
The AUC-ROCs for the best $N$ and $\varepsilon$ using grid-search are reported in Table \ref{table:auc_roc}.
%EXTRASPACE
We see that \DNFP{} achieves near-perfect detection with AUC-ROC of $~99-100\%$ across all attacks.
%EXTRASPACE
On CIFAR-10, across attacks, \DNFP{} outperforms LID on average by \textbf{11.77\%} and KD+BU, KD, BU even more substantially (Table \ref{table:auc_roc}).
%EXTRASPACE
Even compared to LID-white-box (where LID is aware of the attackers mechanism but \DNFP{} is not), \DNFP{} outperforms LID-white-box on average by \textbf{8.0\%} (Appendix, Table \ref{supp:table:whiteboxattack}). 
Further, we find that \DNFP{} is robust across a wide-range of hyperparameter choices. (See Figures \ref{supp:fig:sensitivityAUCROC},\ref{f:multiruns}).
Additionally, we evaluated \DNFP{} against Wasserstein Adversarial Examples, introduced recently in \cite{WAE}. \DNFP{} is able to distinguish between adversarial and clean data with an $98\%+$ AUC-ROC (See Appendix for details).
Wasserstein adversarial examples include perturbations such as scaling, rotation,
translation, and distortion.

\begin{wraptable}{r}{0.4\linewidth}
	\vspace{-0.2in}
	\caption{AUC-ROC of \DNFP{} vs \emph{grey-box attacks} on MiniImagenet-20, $(N,\varepsilon) = (20, 0.05)$. 
		\vspace{-0.05in}
	}
	\label{table:minim}
	\centering
	\begin{small}
		\begin{tabular}{lcc}
			Data & FGSM & BIM-b \\
			\hline
			{MiniImagenet-20}
			& 99.96 & 99.68 
		\end{tabular}
	\end{small}
	\vspace{-0.1in}
\end{wraptable}
\textbf{MiniImagenet-20.}
We test on \emph{MiniImagenet}-20 with $20$ classes randomly chosen (from 100) \citep{NIPS2016_6385} and trained an AlexNet network on 10,600 images (not downsampled) with $91.1\%$  top-1 accuracy.
%EXTRASPACE
We generated test-set adversarial examples using BIM-b with 50 steps \citep{NIPSchallenge} and FGSM.
%EXTRASPACE 
\DNFP{} achieves AUC-ROCs of $>99.5\%$ (Table \ref{table:minim}). 
We could not get results for JSMA and CW-$L_2$, which require too much computation for tasks of this size.
Results for other defenses are not reported due to  unavailable implementations.

A more sophisticated grey-box attack is one where the adversary attempts to use the fingerprint distribution from which the fingerprints are sampled from to mount an adaptive attack. 
We investigate the robustness of \DNFP{} against such an attacker.
For this setup, we randomize both the $\Delta y$s and the $\Delta x$s (See Appendix \ref{supp:sec:greyboxparams} for details).
Several defenses relying on randomness \cite{cisse, xie2018mitigating} have been rendered vulnerable using the expectation-over-transformation (EOT) attack introduced in \cite{obfuscated-gradients}. 
Under this attack, we consider two scenarios: (i) the adversary has access to the exact $\Delta y$ but not $\Delta x$, and (ii) the adversary does not have access to both the $\Delta x$, $\Delta y$.
Note that scenario (i) is not likely in practice, but we evaluate to consider how much information  an attacker needs to render \DNFP{} vulnerable.
In scenario (ii), the adversary attempts to minimize the fingerprint-loss in expectation while producing a misclassification on the original model by minimizing the following objective:
\eq{
	L_{adv}(x,y^*, \theta) + \gamma \mathbb{E}_{\Delta x\sim T_{\Delta x}, \Delta y\sim T_{\Delta y}}\left[ \fploss{}\left(x,y^*,\chi;\theta   
	\right) \right],
	\nonumber
}
where $L_{adv}$  is the misclassification objective, $T_{\Delta x}$ and $T_{\Delta y}$ are the distributions from which $\Delta x$ and $\Delta y$ are sampled respectively. 
To mount the EOT attack, at each iteration, the adversary samples a new set of fingerprints and attempts to minimize the fingerprint-loss while producing a misclassification. 
For this, we use the attack that is successful in the white-box setting described in Appendix \ref{sec:expadapt} for 50000 steps.
Attacks for the other scenarios are constructed similarly. 
The results for these attacks are summarized in Table \ref{table:greyboxEOT}, and we find that the defense is robust against the EOT attack.

%\vspace{-5pt}
\vspace{-5pt}
\section{Conclusion}
\vspace{-5pt}
Our experiments suggest \DNFP{} is efficient at safeguarding DNNs from adversarial attacks.
We find that \DNFP{} is effective at detecting the full range of state-of-the-art adversarial attacks (grey/black-box), 
and the high AUC-ROC scores indicate that the fingerprints generalize well to the test-set, but not to adversarial examples.
%
%We also demonstrate how \DNFP{} can be leveraged for effectively detecting outlier data. 
%
Furthermore, we find that \DNFP{} is robust to a full-range of white-box attacks available in literature; we then develop a new attack to fool \DNFP{}, pushing white-box attacks further.

\bibliography{main}
\bibliographystyle{unsrt}
\clearpage
\pagebreak
\newpage
\begin{appendices}
	\section{Proof for Theorem 1}
	\begin{figure}
		\includegraphics[width=0.7\linewidth]{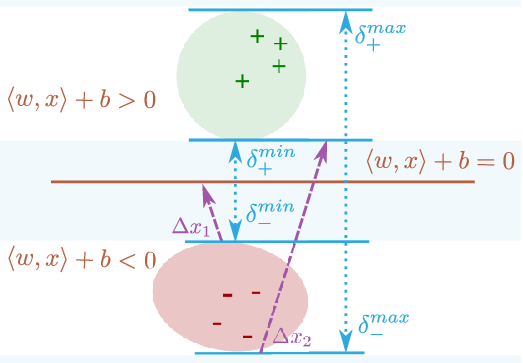}
		\caption{
			Geometry of fingerprints for SVMs with linearly separable data. Let $d(x)$ be the distance of $x$ to the decision boundary (see Thm \ref{thm:svm}). $\delta^{max}_{\pm}$ ($\delta^{min}_{\pm}$) denote the maximal (minimal) distances of the positive ($x_+$) and negative ($x_-$) examples to the separating hyperplane $\brcka{w,x} + b = 0$. The fingerprint $\fpx^{1}$ with $\brcka{\fpx^{1},e} = \delta^{min}_-$ will have $f(x_- + \fpx) < 0$ and $f(x_-) < 0$ for all $x_-$ in the data distribution (red region). Hence, $\fpx^{1}$ will flag all $x'$ in the regions $-\delta^{min}_- < d(x') < 0$ as ``fake'', since for those $x'$ it will \emph{always} see a change in predicted class. Similarly, $\fpx^2$ with $\brcka{\fpx^2,e}=\delta^{max}_-$ always sees a class change for real $x_-$, thus flagging all $x'$ with $d(x') < -\delta^{max}_-$ as~``fake''.
		}
		\label{app:fig:svm}
	\end{figure}
	\begin{proof} Also see Figure \ref{app:fig:svm}.
		Consider any perturbation $\pert{} = \lambda e$ that is positively aligned with $w$, and has $\brcka{\pert{},e} = \delta^{min}_-$. Then for any negative example $(x_-, -1)$ (except for the support vectors that lie exactly $\delta^{min}_-$ from the hyperplane), adding the perturbation $\pert{}$ does not change the class prediction:
		%EXTRASPACE
		\eq{
			%EXTRASPACE
			\sign{f(x_-)} = -1, \hspace{10pt} \sign{f(x_- - \pert{})} = -1.
			%EXTRASPACE
		}
		The fingerprint in \refn{fp3} is an example of such an $\pert{}$.
		%EXTRASPACE
		However, if $\lambda$ is large enough, that is:
		%EXTRASPACE
		\eq{
			\brcka{\pert{}, e} = \delta_{-}^{max},
		}
		(e.g. the fingerprint in \refn{fp4}),
		%EXTRASPACE
		for \textit{all} negative examples $(x_-, -1)$ the class prediction will \emph{always} change (except for the $x_-$ that lie exactly $\delta^{max}_-$ from the hyperplane):
		%EXTRASPACE
		\eq{
			%EXTRASPACE
			\sign{f(x_-)} = -1, \hspace{10pt} \sign{f(x_- + \pert{})} = +1,
			%EXTRASPACE
		}
		%EXTRASPACE
		Note that if $\pert{}$ has a component smaller (or larger) than $\delta^{min}_\pm$, it will exclude \emph{fewer} (more) examples, e.g. those that lie closer to (farther from) the hyperplane.
		%EXTRASPACE
		Similar observations hold for fingerprints \refn{fp1} and \refn{fp2} and the positive examples $x_+$. Hence, it follows that for any $x$ that lies too close to the hyperplane (closer than $\delta^{min}_\pm$), or too far (farther than $\delta^{max}_\pm$), the model output after adding the four fingerprints will never perfectly correspond to their behavior on examples $x$ from the data distribution. For instance, for any $x$ that is closer than $\delta^{min}_+$ to the hyperplane, \refn{fp2} will always cause a change in class, while none was expected. Similar observations hold for the other regions in \refn{svmregions}.
		%EXTRASPACE
		Since the SVM is translation invariant parallel to the hyperplane, the fingerprints can only distinguish examples based on their distance perpendicular to the hyperplane. Hence, this choice of $\lambda$s is optimal.
		%EXTRASPACE
	\end{proof}
	\begin{table}[t]
		\centering
		\begin{tabular}{lc}
			Layer & Parameters \\
			\hline
			Convolution + ReLU + BatchNorm & $11 \times 11 \times 64$\\
			MaxPool & $3 \times 3$  \\
			Convolution + ReLU  + BatchNorm & $5 \times 5 \times 192$  \\
			MaxPool & $3 \times 3$ \\
			Convolution + ReLU + BatchNorm & $3 \times 3 \times 384$ \\
			MaxPool & $3 \times 3$ \\
			Convolution + ReLU  + BatchNorm & $3 \times 3 \times 256$ \\
			MaxPool & $3 \times 3$ \\
			Convolution + ReLU  + BatchNorm & $3 \times 3 \times 156$ \\
			MaxPool & $3 \times 3$  \\
			Fully Connected + ReLU  + BatchNorm& 3072 \\
			Dropout & - \\
			Fully Connected + ReLU + BatchNorm & 1024 \\
			Dropout & - \\
			Softmax & 20 \\
			\\
			\hline
		\end{tabular}
		\caption{MiniImagenet-20 Model Used}
		%EXTRASPACE
		\label{table:miniim_model}
	\end{table}
	
	\begin{table}[t!]
		\centering
		\begin{tabular}{lc}
			Layer & Parameters  \\
			\hline
			Convolution + ReLU + BatchNorm & $5 \times 5 \times 32$ \\
			MaxPool & $2 \times 2$ \\
			Convolution + ReLU  + BatchNorm & $5 \times 5 \times 64$ \\
			MaxPool & $2 \times 2$ \\
			Fully Connected + ReLU  + BatchNorm& 200\\
			Fully Connected + ReLU + BatchNorm & 200\\
			Softmax & 10\\
			\\
			\hline
		\end{tabular}
		\caption{MNIST Model Used}
		%EXTRASPACE
		\label{table:mnist_model}
	\end{table}
	
	\begin{table}[t!]
		\centering
		\begin{tabular}{lc}
			Layer & Parameters  \\
			\hline
			Convolution + ReLU + BatchNorm & $3 \times 3 \times 32$ \\
			Convolution + ReLU  + BatchNorm & $3 \times 3 \times 64$ \\
			MaxPool & $2 \times 2$ \\
			Convolution + ReLU + BatchNorm & $3 \times 3 \times 128$ \\
			Convolution + ReLU  + BatchNorm & $3 \times 3 \times 128$ \\
			MaxPool & $2 \times 2$ \\
			Fully Connected + ReLU  + BatchNorm& 256\\
			Fully Connected + ReLU + BatchNorm & 256\\
			Softmax & 10\\
			\\
			\hline
		\end{tabular}
		\caption{CIFAR Model Used}
		%EXTRASPACE
		\label{supp:table:cifar_model}
	\end{table}

	\section{Models for Evaluation}
	\label{app:app:sec:appendix}
	Note: Code for CW-adaptive is based on code from \url{https://github.com/carlini/nn_robust_attacks}. %EXTRASPACE
	We use published code from \citep{WAE} for Wasserstein Adversarial Attack, and the other attacks were obtained from the
	paper \citep{ma2018characterizing}.
	
	\subsection{MNIST}
	For MNIST, we use the model described in Table \ref{table:mnist_model}.
	
	\subsection{CIFAR-10}
	For CIFAR-10, we use the model described in Table \ref{supp:table:cifar_model}
	
	\subsection{MiniImagenet-20}
	\paragraph{MiniAlexNet Model}
	We use a model similar to AlexNet for MiniImagenet-20. The model used is described in
	Table \ref{table:miniim_model}
	
	\paragraph{MiniImagenet-20 classes}
	We use images from the following 20 ImageNet classes for our experiments:\\
	\texttt{n01558993,  n02795169,  n03062245,  n03272010,  n03980874,  n04515003
		n02110063,  n02950826,  n03146219,  n03400231,  n04146614,  n04612504,
		n02443484,  n02981792,  n03207743,  n03476684,  n04443257,  n07697537}
	
	% MNIST FGSM 
	%

	\section{Grey-box Evaluation -- $\Delta y$ Hyperparameters}
	\label{supp:sec:greyboxparams}
	Instead of simple $\Delta y^{i,j}$, we can encode more complex fingerprints that are harder to guess for an adversary.
	For the grey-box threat-model evaluation in Section \ref{sec:effective}, 
	for each $\Delta x^i$, we sample $\Delta y^{i,j} \in \mathbb{R}^{10}$ such that each entry of $\Delta y^{i,j}$ is uniformly drawn from $[-0.5,0.5]$.
	This random sampling minimizes the structural assumptions the adversary can make about the fingerprints.
	
	\section{Sensitivity analysis}
\begin{figure}[H]
		\centering \includegraphics[width = 0.32\textwidth]{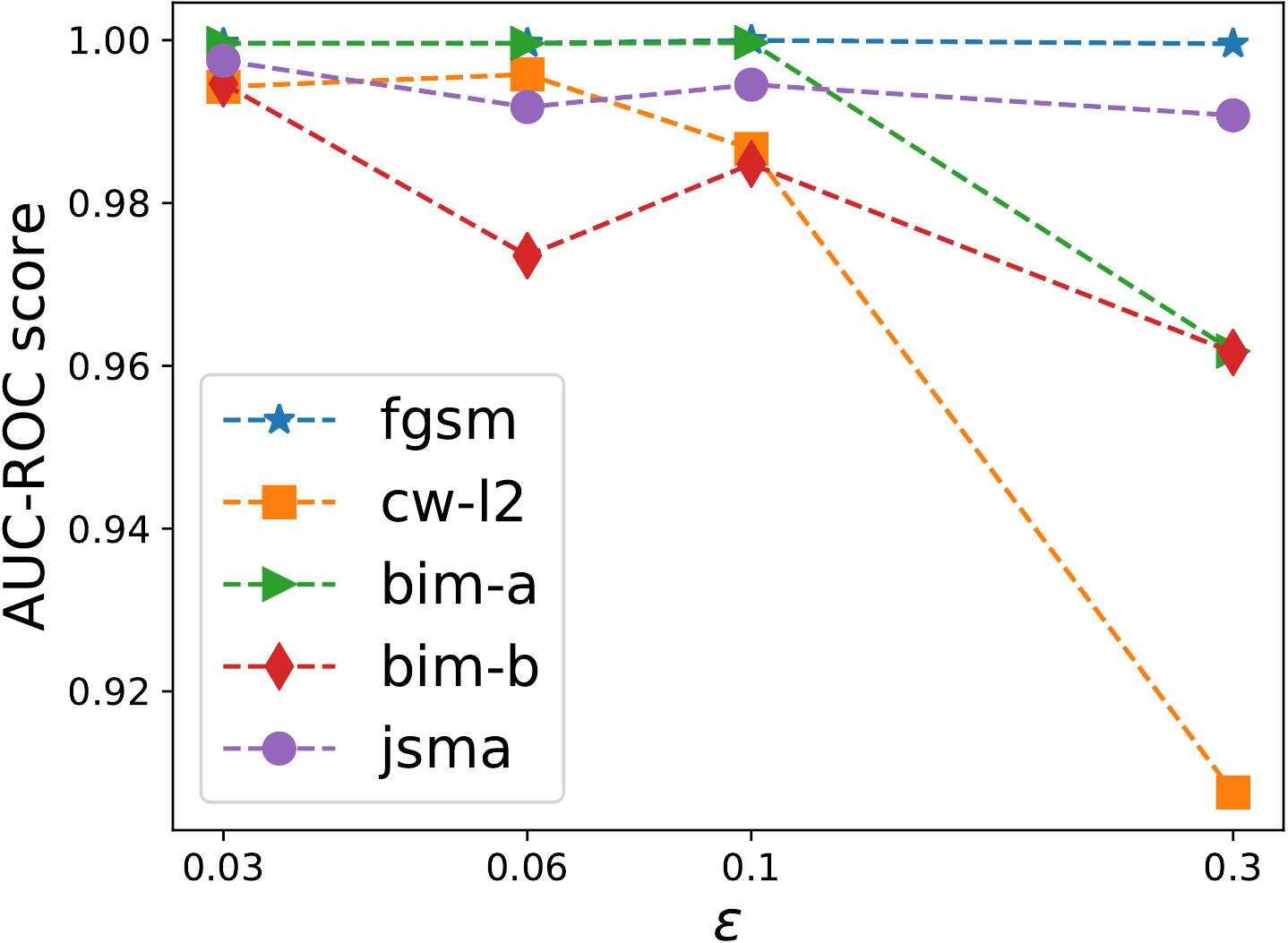}
		\centering \includegraphics[width = 0.32\textwidth]{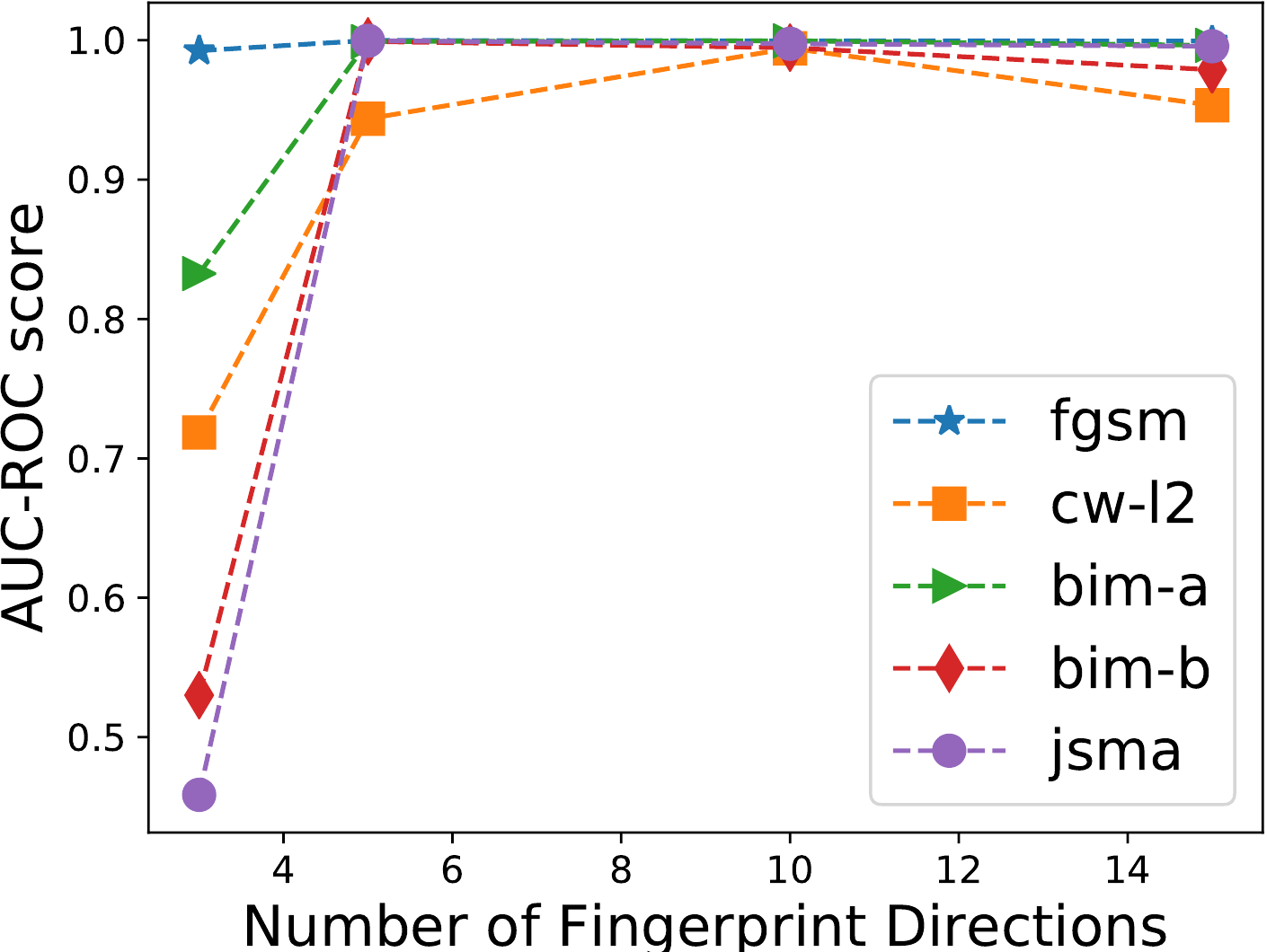}
		\centering \includegraphics[width = 0.32\textwidth]{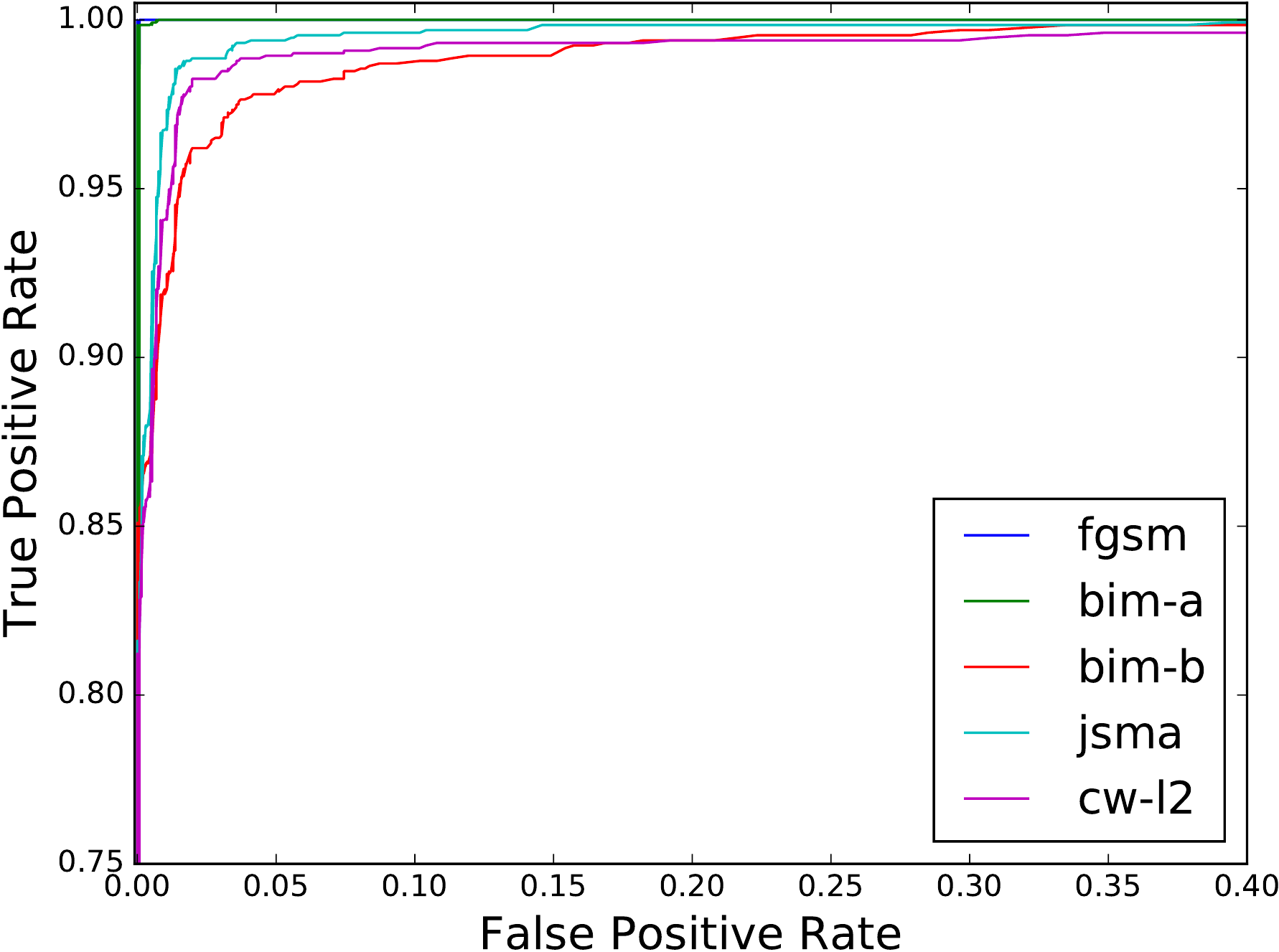}
		\caption{AUC-ROC performance for different hyperparameter settings (top-left: Varying $\varepsilon$ ($N=10$), top-right:Varying $N$ ($\varepsilon=0.03$)) and ROC curves (bottom) on MNIST. We see that the performance of \DNFP{} is robust across attacks and hyperparameters, with the AUC-ROC between $90-100\%$ for most settings. The AUC-ROC is lowest versus CW-$L_2$, which is one of the strongest known attack.}
		\label{supp:fig:sensitivityAUCROC}
	\end{figure}

\begin{figure}[t]
		\includegraphics[width = 0.32 \textwidth]{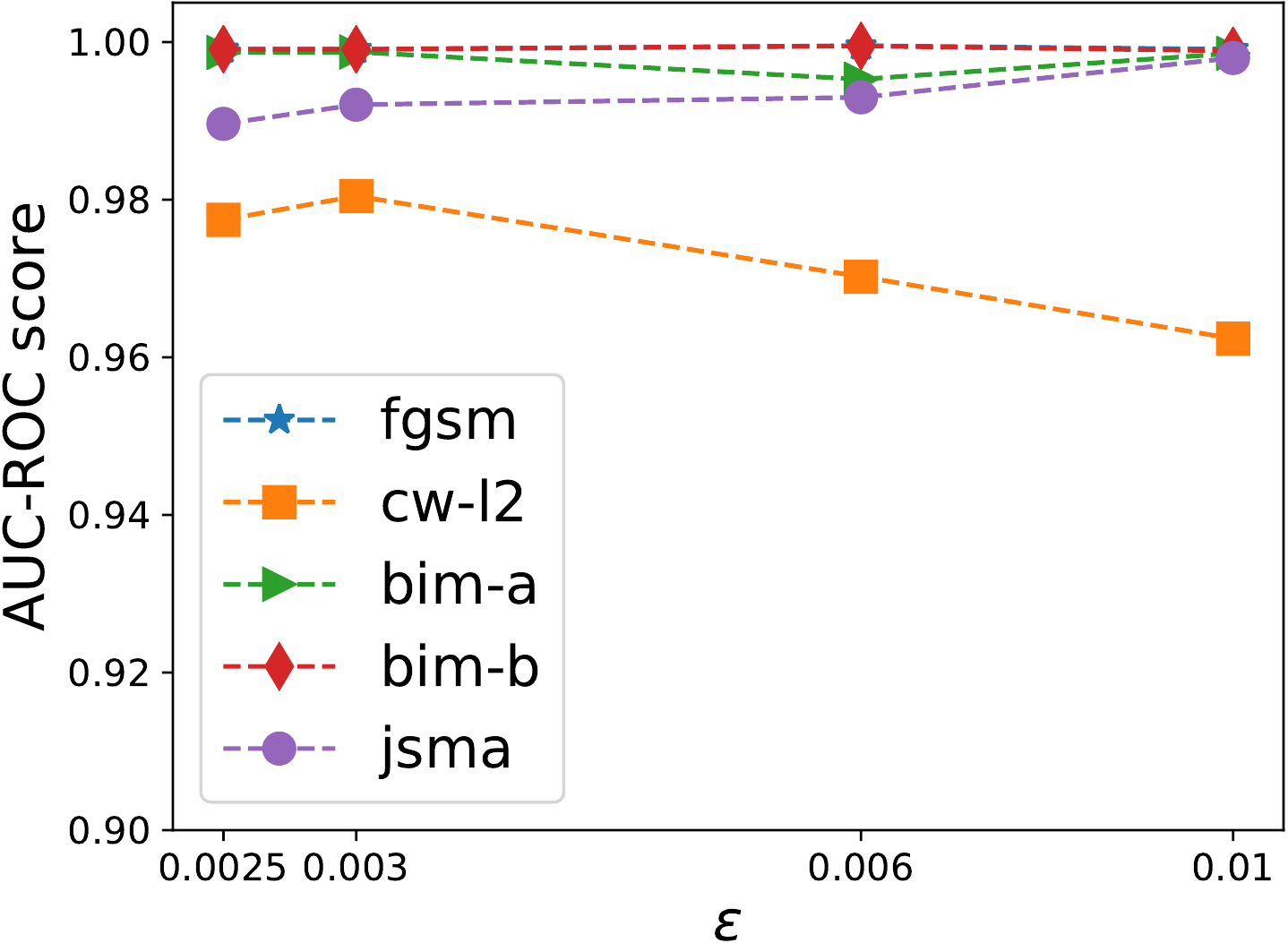}
		\includegraphics[width = 0.32 \textwidth]{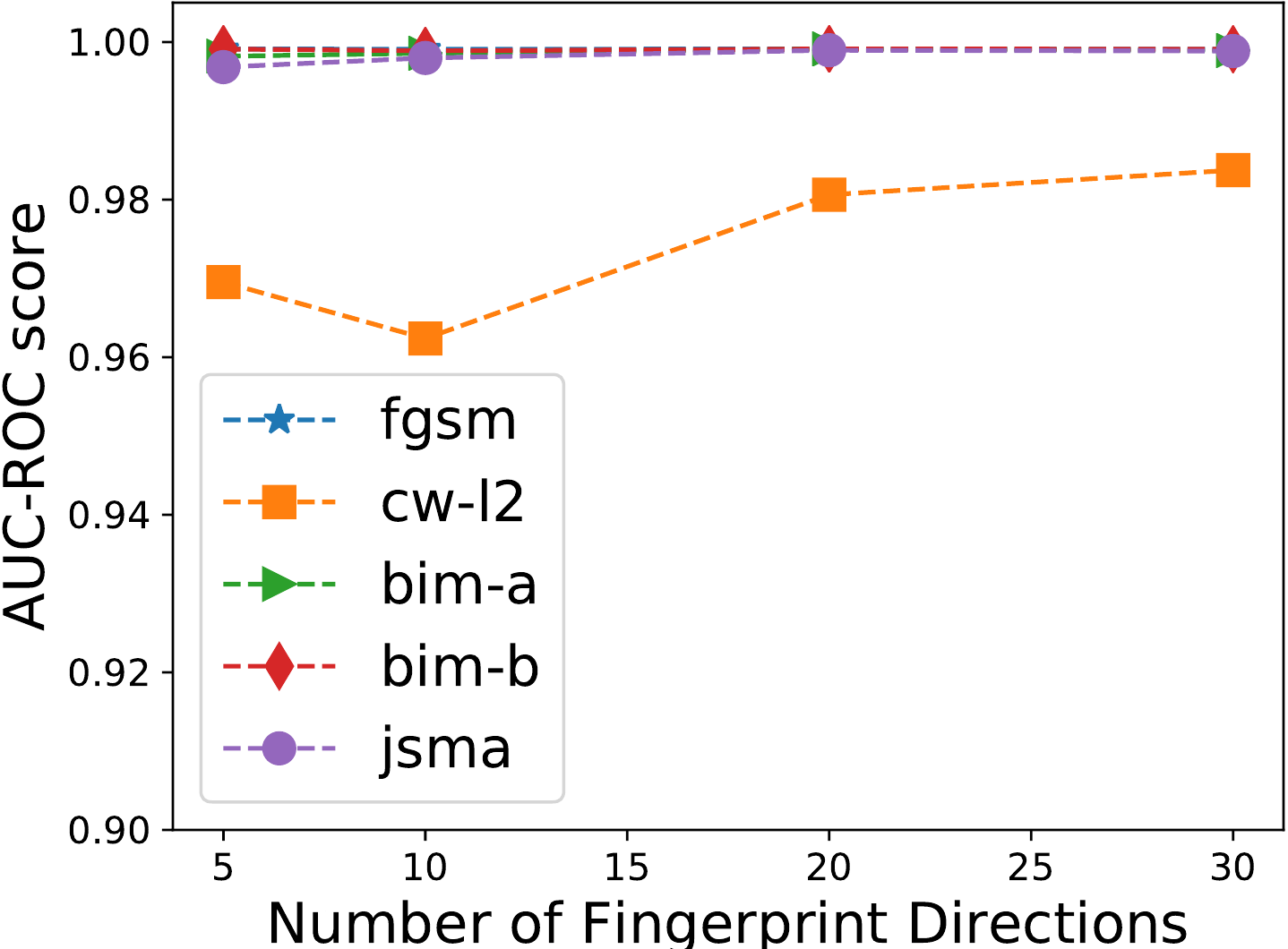}
		\includegraphics[width = 0.32\textwidth]{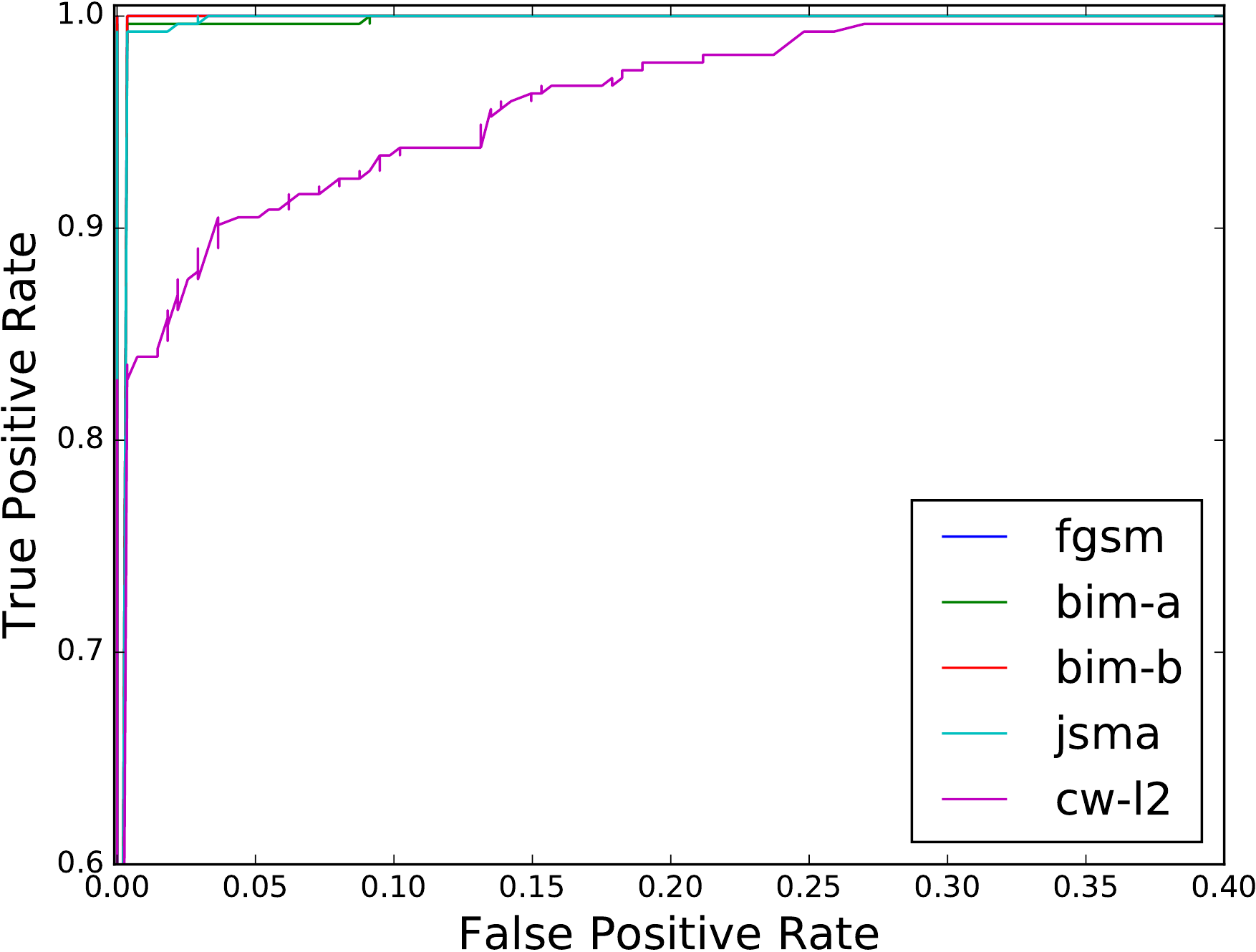}
		\caption{AUC-ROC for different hyperparameters (top-left:{Varying fingerprint magnitude $\varepsilon$ {($N=10$)}}, top-right: {Varying no. fingerprints ($N$) ($\varepsilon=0.01$)}) and ROC curves (bottom) on CIFAR-10 for partial-whitebox attacks. For analysis on MNIST, see Appendix. \DNFP{} is robust across attacks \& hyperparameters with an AUC-ROC between $95-100\%$. Increasing $N$ improves performance, indicating more fingerprints are harder to fool. Increasing the magnitude $\varepsilon$ decreases AUC on CW-$L_2$ only, suggesting that as adversarial perturbations become of smaller magnitude, \DNFP{} requires smaller $\varepsilon$.}
		\label{fig:sensitivityAUCROC}
	\end{figure}

	We study the effect of changing $N$ (number of fingerprint directions) and $\varepsilon$ (magnitude of fingerprint-perturbation $\Delta x$) on the AUC-ROC for CIFAR-10 and MNIST.
	%EXTRASPACE
	Figure \ref{fig:sensitivityAUCROC} and \ref{supp:fig:sensitivityAUCROC} show that \DNFP{} performs well across a wide range of hyperparameters and is robust to variation in the hyperparameters for PWAs.
	%EXTRASPACE
	With increasing $\varepsilon$, the AUC-ROC for CW-${L_2}$ decreases. As discussed before, a possible explanation is that CW-${L_2}$ produces smaller adversarial perturbations than other attacks, and for larger fingerprint-distortions $\varepsilon$, the fingerprints are less sensitive to those small adversarial perturbations.
	%EXTRASPACE
	However, the degradation in performance is not substantial ($\sim 4-8\%$) as we increase $\varepsilon$ over an order of magnitude.
	%EXTRASPACE
	With increasing $N$, the AUC-ROC generally increases across attacks.
	%EXTRASPACE
	We conjecture that larger sets of fingerprints can detect perturbations in more directions and results in better detection.
	
\begin{figure}
		\centering
		\includegraphics[width=0.6\linewidth]{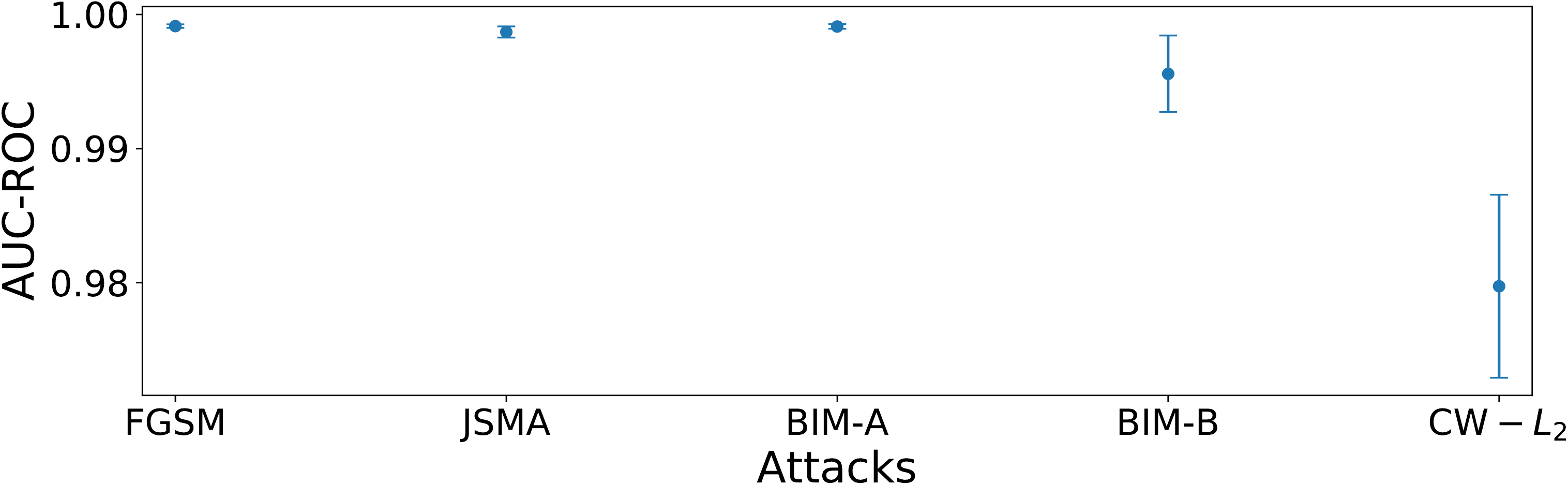}
		\caption{AUC-ROC mean $\mu$ and standard-deviation $\sigma$ for 32 randomly sampled fingerprints (including randomizing $N$) for CIFAR-10. The AUC-ROC across all PWAs varies little ($\sigma<1\%$), with $\sigma$ highest for CW-${L_2}$.
		}\label{f:multiruns}
	\end{figure}

	Figure \ref{f:multiruns} shows that \DNFP{} achieves mean AUC-ROC of $98\%-100\%$  against all PWA, with standard deviation $<1\%$.
	%
	%EXTRASPACE
	This suggests that \DNFP{} is \emph{not very sensitive to the chosen fingerprints}.
	%EXTRASPACE
	%\
	\section{Randomized Queries -- sanity check for query based attacks}
	\label{supp:sec:randomsampling}
\begin{figure}[t]
		\centering
		\includegraphics[width = 0.43\textwidth]{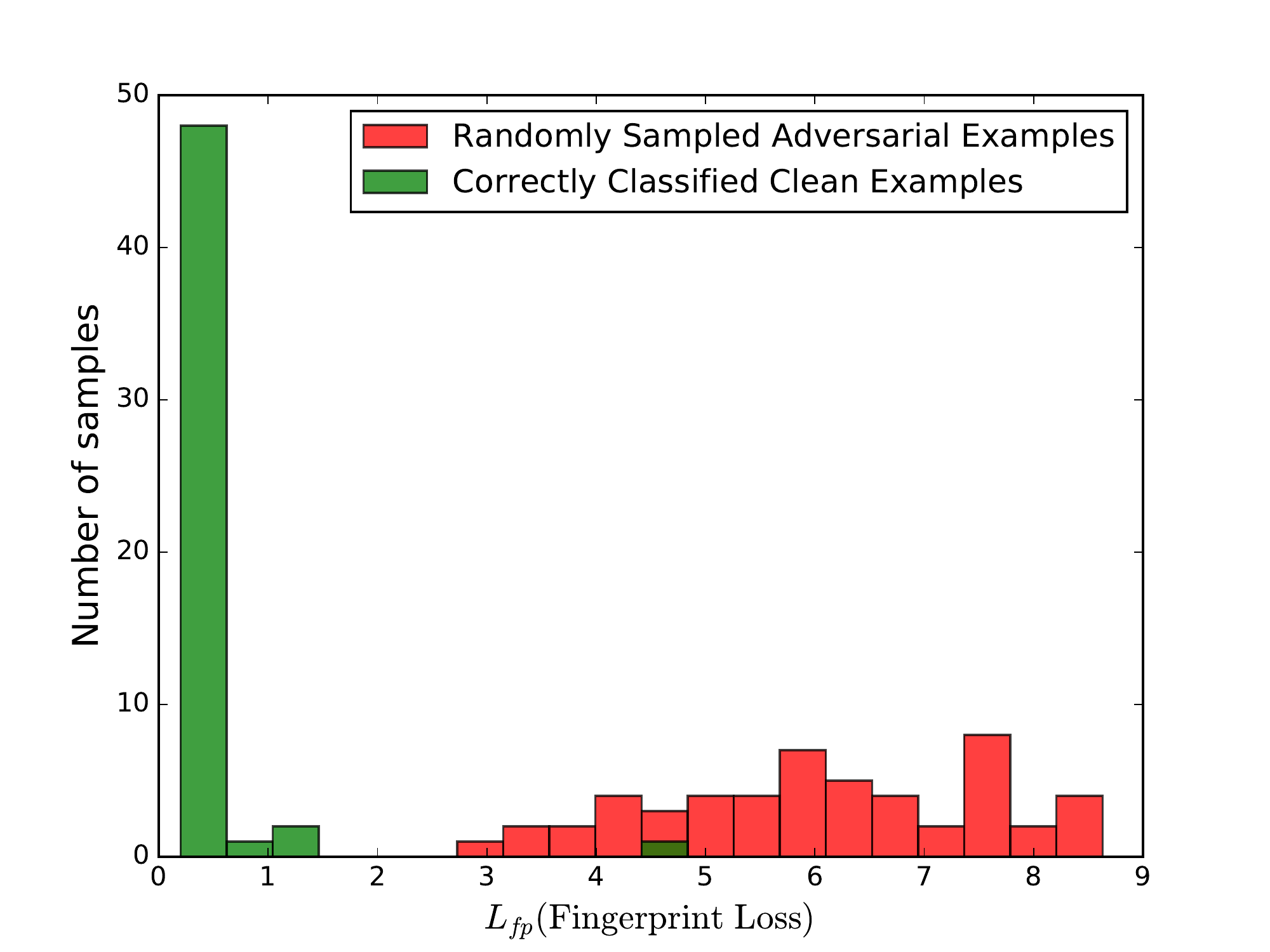}
		\includegraphics[width = 0.43\textwidth]{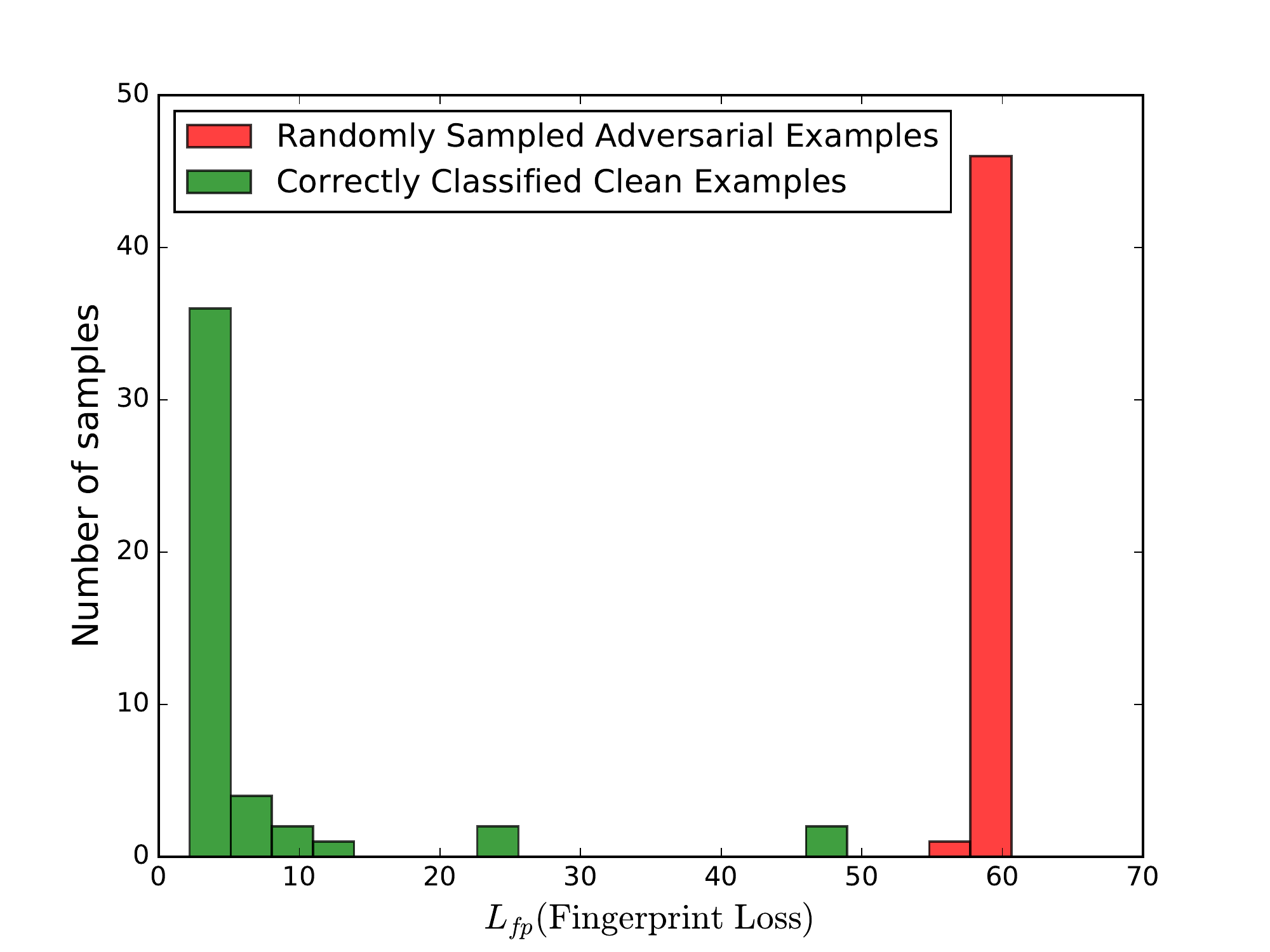}
		\caption{Top: MNIST, Bottom: CIFAR-10. Histograms depicting the distribution of losses for randomly sampled adversarial examples and test-data.
			Randomly sampled adversarial examples are well separated from unseen test examples.
			For \DNFP{}, hyperparameters are $(\varepsilon,N)=(0.1,10)$ and $(\varepsilon,N)=(0.003,30)$
			for MNIST and CIFAR-10 respectively. }
		\label{supp:fig:randomsampling}
	\end{figure}

	To evaluate the feasibility of query based black-box attacks such as \citep{nesblackbox, decisionbased}, we sample randomly in $\epsilon$-balls around unseen test-data to find adversarial examples. 
	For 100 MNIST test samples, we sample $10^6$ points uniformly from the $l_\infty$-ball with $\lVert \eta \rVert_\infty \leq 0.3$ and for 50 CIFAR-10 test samples, we sample $5 \times 10^5$ points uniformly from the $l_\infty$-ball with $\lVert \eta \rVert_\infty \leq 0.25$.
	Amongst these randomly sampled points for each test point, we compute the smallest $\underset{i}{\min} \left(L_{fp}\left(x,i,\xi;\theta\right)\right)$ (let us call this value $L_{fp}^*$) across points that are adversarial (i.e. cause a misclassification).
	Note that $L_{fp}^*$ for each test point is the minimum fingerprint-loss over all adversarial samples sampled around the point and across all labels.
	For CIFAR-10, we are able to find adversarial examples for 47 points and for MNIST, we are able to find adversarial examples for 52 points.
	Figure \ref{supp:fig:randomsampling} shows the distribution of $L_{fp}$ across the test points and $L_{fp}^*$ across the corresponding randomly sampled adversarial points.
	We observe that for MNIST, one single test sample has a high fingerprint-loss while all adversarial samples have high fingerprint-losses. 
	The remaining test-samples have fingerprint-losses that are roughly $3-30$ times smaller than the adversarial examples.
	For CIFAR-10, the test and adversarial points are well separated, with most test samples having losses significantly smaller (roughly $10$ times smaller) than the randomly sampled adversarial examples.
	This indicates that it is extremely difficult to perform query based black-box adversarial attacks against \DNFP{}.
	This also indicates that \DNFP{} removes most of the adversarial examples, and does not simply gradient mask. 
	\section{Whitebox Analysis of \DNFP{}} 
	\label{sec:expadapt}
	The strongest threat-model, whitebox-attack, is one where the adversary has access to the parameters of \DNFP{}.
	To evaluate whether \DNFP{} is robust in this setting, ), we consider adaptive variants of FGSM, BIM-b, CW-${L_2}$, and SPSA \citep{obscurityicml}.
	Under this threat-model the attacker tries to find an adversarial example $x'$ that also minimizes the fingerprint-loss \eqref{eq:encode}, while attacking the model trained with \DNFP{}.
	We find that \DNFP{} is robust across the full range of such state-of-the-art attacks, achieving AUC-ROCs of 96-100\% (See Table \ref{table:adaptive_auc_roc})
	
	\paragraph{Adaptive-FGSM, Adaptive-BIM-b, Adaptive-SPSA}
	For the FGSM, BIM-b and SPSA (untargeted) attacks we mount an adaptive attack with a modified optimization objective as in \citep{obscurityicml}. Specifically, for SPSA, the loss function to minimize is:
	\eq{J_{\theta}^\mathrm{adv}(x',y^*, \theta) + \gamma \fploss{}\left(x',y^*,\chi;\theta
		\right),
		\nonumber
	}
	where $J_{\theta}^\mathrm{adv}$ is the original adversarial objective from \citep{obscurityicml}.
	For the gradient-based FGSM and BIM-B attacks, we use gradients of the following loss function:
	\eq{L_{CE}(x,y^*, \theta) - \gamma \fploss{}\left(x,y^*,\chi;\theta
		\right),
		\nonumber
	}
	where $L_{CE}(x,y^*)$ is the cross-entropy loss.
	For each of the attacks and for each data-point, we choose the largest $\gamma\in[10^-3,10^4]$ that results in a successful attack 
	with a bisection search over $\gamma$ -- note that larger $\gamma$ values increase the priority for minimizing $L_{fp}$. 
	For the three adaptive attacks, the perturbation bounds are $\lVert \eta\rVert_\infty \leq 0.4$ for MNIST and $\lVert \eta\rVert_\infty \leq 0.05$ for CIFAR-10.

	\paragraph{Adaptive-CW-${L_2}$}
	We consider two adaptive variants of the CW-${L_2}$ attack.
	The first variant we consider is with the modified objective function:
	\eq{
		%EXTRASPACE
		\underset{x'}\min\,\,{\| x-x'\|_2 +
			\gamma_1 \left(L_{\textrm{CW}}\left(x'\right)+ \gamma_2 \fploss{}\left(x',y^*,\chi;\theta
			\right)\right)} .
		%EXTRASPACE
	}
	Here, $y^*$ is the label-vector,
	%EXTRASPACE
	$\gamma_1 \in [10^{-3},10^6]$ and $\gamma_2 \in [10^{-3},10^4]$ are scalar coefficients,
	%EXTRASPACE
	$\fploss{}$ is the fingerprint-loss we trained on
	%EXTRASPACE
	and $L_{\textrm{CW}}$ is an objective encouraging misclassification.
	%EXTRASPACE
	%
	To find $\gamma_1$ and $\gamma_2$ we do a bisection search, first decreasing $\gamma_1$ (as in \cite{CarliniBreak}) and then increasing $\gamma_2$ gradually in a similar manner. 
	Note that increasing $\gamma_2$ increases the importance given to minimizing $\fploss{}$.
	The successful attack with the smallest $\fploss{}$ during our search is chosen.
	
	The second variant is similar to the one considered in \citep{CarliniBreak}.
	Here $\gamma_2$ is held at $1.0$ and the successful attack  with the smallest $\| x-x'\|_2$ is chosen during a bisection search over $\gamma_1$.
	
	%EXTRASPACE
	%
	\begin{comment}
	Under this threat model, \DNFP{} achieves an AUC-ROC of \textbf{98.79\%} for the CIFAR-10 task  against Adaptive-CW-${L_2}$, with $N=30$ and $\varepsilon=0.006$ for a set of unseen test-samples (1024 \emph{pre-test}) and the corresponding adversarial examples.
	%EXTRASPACE
	In contrast to other defenses vulnerable
	to Adaptive-CW-${L_2}$ \citep{CarliniBreak}, we find that \DNFP{} is robust even under this adaptive whitebox-attack threat model.
	%
	\end{comment}
	\begin{table*}[t!]
		\centering
		\resizebox{\textwidth}{!}{
			\begin{tabular}{llccccc}
				Data & Method & Adaptive-FGSM &  Adaptive-BIM-b & Adaptive-CW-${L_2}$ & Adaptive-CW-${L_2}$ ($\gamma_2=1$) & Adaptive-SPSA  \\
				\hline
				\\
				{MNIST}
				& \DNFP{}  & {99.91}  & {99.37} & {95.04} & {99.17} & {99.94}\\
				{CIFAR-10}
				& \DNFP{}  & {99.99} & {99.92} & {97.19} & {97.56} & {99.99}\\
		\end{tabular}}
		\vspace{-5pt}
		\caption{Detection AUC-ROC for adaptive whitebox attacks on datasets MNIST and CIFAR-10. 
			Other defenses such as \citep{GAN1,guideddenoising}, including the baselines KD and BU, fail under adaptive-attacks ($<10\%$ accuracy).
			For MNIST, the \DNFP{}  parameters for FGSM, SPSA are $(\varepsilon,N)=(0.1,10)$ and $(\varepsilon,N)=(0.05,20)$ for the BIM-b, CW-$L_2$ attacks.
			For CIFAR-10, the parameters are set at  $(\varepsilon,N)=(0.003,30)$ across attacks.
			% 
			%\CORR{Note that whitebox LID detection is unfair comparison, because \DNFP{} is blackbox-defense.}
			\vspace{-10pt}
		}
		%However, LID only outperforms \DNFP{} slightly in the whitebox setting, i.e. with data augmentation with CW-${L_2}$ adversarial examples, while \DNFP{} is always attack agnostic. On MiniImagenet-20, \DNFP{} achieves near-perfect AUC-ROC versus FGSM and BIM-b. Note, for MiniImagenet-20, 1) attacks are prohibitively slow (JSMA, CW-$L_2$) and 2) due to time constraints, LID scores were not obtainable.
		%EXTRASPACE
		\label{table:adaptive_auc_roc}
	\end{table*}
	
	\subsection{New White-Box Attack for Breaking \DNFP{}}
	While \DNFP{} is robust to the full range of attacks studied above, we find that this is because of the highly nonconvex problem the attacker is required to solve.
	During detailed analysis we notice that during the multi-objective optimization of minimizing the fingerprint loss($L_{fp}$) and the misclassification loss($L_{adv}$), the individual losses oscillate, which makes it difficult to craft adversarial examples with low fingerprint loss.
	To solve this problem, we consider a new attack that can adaptively solve this optimization problem stabilizing the two losses and generating adversarial examples that successfully fool \DNFP{}. 

	Unlike current adaptive attacks which use a fixed coefficient $(\gamma)$ when attacking detection models with multiple loss(e.g.: $L_{fp}$ and $L_{adv}$ in \DNFP{}), we scale both losses to balance the gradients from the two losses and simultaneously optimize both of them. 
	To do this, we introduce two functions that vary monotonically as a function of the difference between the fingerprint-loss of the adversarial sample ($x'$) and the real sample ($x$).
	Define $\Delta L_{fp}(x,x')= \fploss{}(x',y',\fp{};\theta) - \fploss{}(x,y,\fp{};\theta)$, where $y'=f(x')$.
	Define two functions:
	\[f_1(x',x):=10^{\Delta L_{fp}(x,x')},\]
	and
	\[f_2(x',x):=0.1^{\Delta L_{fp}(x,x')},\]
	We then perform an optimization based attack (with SGD) to minimize:
	\[
	f_1(x',x) \fploss{}(x',y',\fp{};\theta) + L_{adv} f_2(x',x).
	\]
	We find that this cost function effectively optimizes both losses simultaneously, both producing a misclassification and minimizing the fingerprint-loss, thereby fooling the detector.
	Using this attack (50000 steps with learning-rate=0.00001) with a distortion bound of $\lVert \eta_\infty \rVert \leq 0.031$, we are able to reduce the AUC-ROC to about 0.5.
	This implies that the adversarial examples are virtually indistinguishable from real samples.
	We use this successful whitebox-attack to evaluate EOT and transfer based attacks in the grey-box/black-box setting.

	\begin{table}[t]
		\centering
		\begin{tabular}{llccccc}
			Data & Method & FGM & JSMA & BIM-a & BIM-b & CW-${L_2}$ \\
			\hline
			\multirow{2}{*}{MNIST}
			& LID &  99.68 & 98.67 & 99.61 & 99.90 & 99.55 \\
			& \DNFP{}  & \bf{100.0} & \bf{99.97} & \bf{99.94} & \bf{99.98} & \bf{99.74} \\
			\hline
			\multirow{2}{*}{CIFAR-10}
			& LID &  82.38 & 95.87 & 82.30 & 99.78 & \bf{98.94} \\
			& \DNFP{}  & \bf{99.96} & \bf{99.91} & \bf{99.91} & \bf{99.95} & {98.87} \\
		\end{tabular}

		\caption{Detection AUC-ROC for \DNFP{},whitebox-LID against \emph{whitebox-attackers} (know model $f(x;\theta)$, but not fingerprints; see Section \ref{sec:threatmodel}), on MNIST, CIFAR-10 tasks on test-set (``real'') and corresponding adversarial (``fake'') samples (1328 \emph{pre-test} samples each). \DNFP{} outperforms the baselines (LID, KD, BU) on MNIST and CIFAR-10 across all attacks, except CW-L2 where it performs comparably.
			A possibly explanation for LID's improved performance against stronger, iterative attacks is gradient masking \cite{obfuscated-gradients}. 
		}
		\label{supp:table:whiteboxattack}
		%EXTRASPACE
	\end{table}
	
	\section{Attack Hyperparameters}
	Tables \ref{supp:table:hyperparam}, \ref{table:attackparam} summarize the hyperparameter settings corresponding to the experiments in Tables \ref{table:auc_roc},\ref{table:adaptive_auc_roc}, and \ref{table:wasserstein}. 
	\begin{table}[t]
		\caption{Attack hyperparameters corresponding to Tables \ref{table:auc_roc},\ref{table:adaptive_auc_roc}, and \ref{table:wasserstein}.
			%
			%\CORR{Note that whitebox LID detection is unfair comparison, because \DNFP{} is blackbox-defense.}
		}
		\centering
		\small   
		\begin{tabular}{lllllllc}
			Method & Parameters \\
			\hline
			{CW-$L-2$} & Bisection-steps ($\gamma_1$)=9, Bisection-steps ($\gamma_2$)=5, Max Iteration Steps = 1000, 
			\\ & L2\_ABORT\_EARLY = True,  L2\_LEARNING\_RATE = 1e-2, L2\_TARGETED = True,
			\\ &   L2\_CONFIDENCE = 0, L2\_INITIAL\_CONST = 1e-3    ($\gamma_1$ initial value)
			\\ & L2\_INITIAL\_CONST\_2 = 0.1   ($\gamma_2$ initial value)   \\
			{SPSA} & Bisection steps=6, Upper-bound for bisection = 50.0, \\
			& spsa-iters=1, spsa-samples=128, Lower-bound=0.001 \\
			& Iterations = 100, lr=0.01,dr=0.01\\
			{BIM (MNIST)} & eps-iter=0.010\\
			{BIM (CIFAR)} & eps-iter=0.005\\
			{Wasserstein (CIFAR)} & Regularization=3000, Wasserstein-distance(p)=2, PGD Step Size(alpha)=0.1, \\ & epsilon=0.01, epsilon\_factor=1.17, maxiters=400\\
		\end{tabular}
		%However, LID only outperforms \DNFP{} slightly in the whitebox setting, i.e. with data augmentation with CW-${L_2}$ adversarial examples, while \DNFP{} is always attack agnostic. On MiniImagenet-20, \DNFP{} achieves near-perfect AUC-ROC versus FGM and BIM-b. Note, for MiniImagenet-20, 1) attacks are prohibitively slow (JSMA, CW-$L_2$) and 2) due to time constraints, LID scores were not obtainable.
		%EXTRASPACE
		\label{supp:table:hyperparam}
	\end{table}
	\begin{table*}[t]
		\caption{Parameters and model test-accuracy on PW-attacks for different datasets (without \DNFP{} test). CW-$L_2$, Wasserstein and JSMA attacks are unbounded. The bounds are relative to images with pixel intensities in [0, 1].}
		\begin{tabular}{llccc}
			Data &  Attack &  Test  & Bound on Adversarial\\
			& & Accuracy & Perturbation $\eta$ \\
			\hline
			\multirow{2}{*}{MNIST}
			& FGSM & 11.87\% & $\norm{\eta}_\infty \leq$ 0.4\\
			& BIM-a & 0.00\%& $\norm{\eta}_\infty \leq$ 0.4\\
			& BIM-b & 0.00\%& $\norm{\eta}_\infty \leq$ 0.4\\
			& JSMA & 1.73\% \\
			& CW-$L_2$ & 0.00\% \\
			\hline
			\multirow{2}{*}{CIFAR} & FGSM & 11.39 \% & $\norm{\eta}_\infty \leq$ 0.05 \\
			& BIM-a & 0.00\% & $\norm{\eta}_\infty \leq$ 0.05 \\
			& BIM-b & 0.00\% & $\norm{\eta}_\infty \leq$ 0.05\\
			& JSMA & 13.33\% \\
			& CW-$L_2$ & 0.00\% \\
			& Wasserstein & 9.08\% \\
			\hline
			\multirow{2}{*}{MiniImagenet} & FGSM & 0.00\% &  $\norm{\eta}_\infty \leq$ 16/255\\
			& BIM-b & 0.00\% &  $\norm{\eta}_\infty \leq$ 16/255\\
		\end{tabular}
		\label{table:attackparam}
	\end{table*}
	
	\begin{table}[t]
		\caption{Detection AUC-ROC for \DNFP{} against greybox Wasserstein adversarial attack on CIFAR-10 task.}
		\begin{tabular}{llcc}
			Data & Attack & AUC-ROC \\
			\hline
			{CIFAR-10} & Wasserstein & 98.14\\
		\end{tabular}
		\label{table:wasserstein}
	\end{table}
	\section{Wasserstein Adversarial Example}
	\DNFP{} is able to distingish between Wasserstein Adversarial Example and clean-data efficiently (Table \ref{table:wasserstein}). We increase the default epsilon value from 0.001 to 0.01 to achieve more than 90\% attack success rate. The remaining hyperparameters are the same as in \citep{WAE}.

\end{appendices}

\end{document}